\documentclass[10pt,twocolumn,letterpaper]{article}

\usepackage{cvpr}              

\usepackage{algorithm,algorithmicx,algpseudocode}

\newcommand{\bI}{\mathbf{I}}

\newcommand{\bzero}{\mathbf{0}}

\newcommand{\bz}{\mathbf{z}}

\newcommand{\bepsilon}{{\boldsymbol{\epsilon}}}

\definecolor{cvprblue}{rgb}{0.21,0.49,0.74}
\usepackage[pagebackref,breaklinks,colorlinks,allcolors=cvprblue]{hyperref}
\usepackage{multirow, adjustbox}

\def\paperID{3088} 
\def\confName{CVPR}
\def\confYear{2025}

\title{Learning Hazing to Dehazing: Towards Realistic Haze Generation for Real-World Image Dehazing}

\author{Ruiyi Wang$^1$ \quad   Yushuo Zheng$^1$ \quad   Zicheng Zhang$^1$ \quad   Chunyi Li$^1$ \quad Shuaicheng Liu$^2$ \\  Guangtao Zhai$^1$ \quad  \stepcounter{footnote}Xiaohong Liu$^1$\thanks{~Corresponding author.}\vspace{2mm}\\
$^1$Shanghai Jiao Tong University\quad\ $^2$University of Electronic Science and Technology of China\\
}

\begin{document}
\maketitle
\begin{abstract}
\hspace*{\parindent}Existing real-world image dehazing methods primarily attempt to fine-tune pre-trained models or adapt their inference procedures, thus heavily relying on the pre-trained models and associated training data. Moreover, restoring heavily distorted information under dense haze requires generative diffusion models, whose potential in dehazing remains underutilized partly due to their lengthy sampling processes. To address these limitations, we introduce a novel hazing-dehazing pipeline consisting of a Realistic Hazy Image Generation framework (HazeGen) and a Diffusion-based Dehazing framework (DiffDehaze). Specifically, HazeGen harnesses robust generative diffusion priors of real-world hazy images embedded in a pre-trained text-to-image diffusion model. By employing specialized hybrid training and blended sampling strategies, HazeGen produces realistic and diverse hazy images as high-quality training data for DiffDehaze. To alleviate the inefficiency and fidelity concerns associated with diffusion-based methods, DiffDehaze adopts an Accelerated Fidelity-Preserving Sampling process (AccSamp). The core of AccSamp is the Tiled Statistical Alignment Operation (AlignOp), which can provide a clean and faithful dehazing estimate within a small fraction of sampling steps to reduce complexity and enable effective fidelity guidance. Extensive experiments demonstrate the superior dehazing performance and visual quality of our approach over existing methods. The code is available at \url{https://github.com/ruiyi-w/Learning-Hazing-to-Dehazing}.
\end{abstract}    
\section{Introduction}

\begin{figure}[t]
  \centering
  \includegraphics[width=\linewidth]{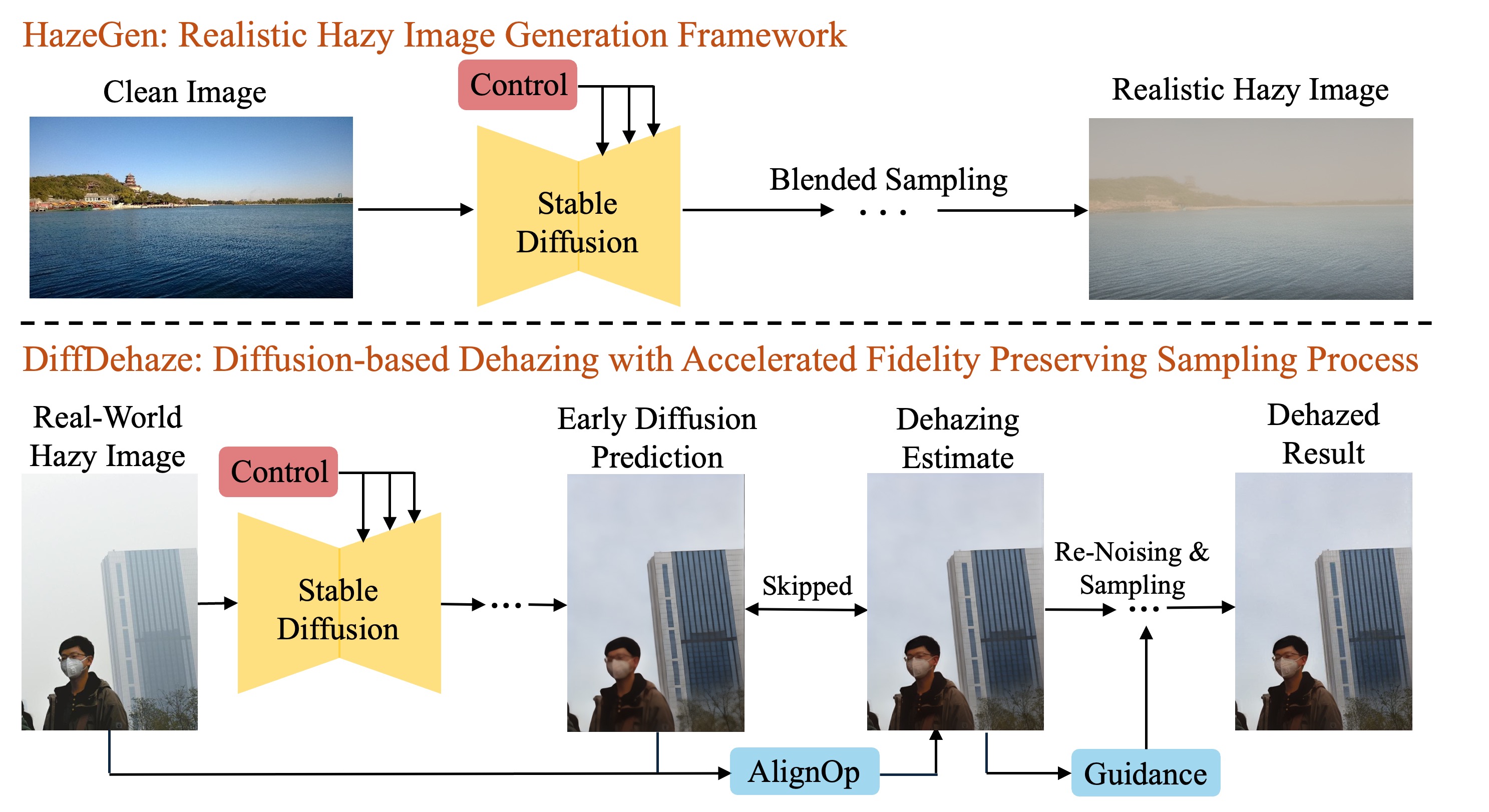}
  \caption{Overview of the proposed pipeline. HazeGen utilizes a pre-trained text-to-image diffusion model to generate realistic hazy images, which serve as the training data for DiffDehaze. DiffDehaze adopts an Accelerated Fidelity-Preserving Sampling process (AccSamp) that effectively reduces sampling steps while producing superior dehazing results with enhanced fidelity.
}
  \label{fig:flow_chart}
\end{figure}
\hspace*{\parindent} Images captured under hazy conditions frequently exhibit color distortion and detail loss, significantly impairing the performance of downstream vision tasks. The formation of hazy images is typically modeled by the physical scattering model~\cite{asm1, asm2}:
\begin{equation}
  I(x) = J(x)\cdot t(x)+A\cdot (1-t(x)),
  \label{eq:asm}
\end{equation}
where $I(x)$ represents a hazy image and $J(x)$ is its clean counterpart. The parameters $A$ and $t(x)$ denote the global atmospheric light and the transmission map, respectively.

Estimating the parameters, particularly the transmission map, from a single hazy image is inherently ill-posed. Early methods incorporated statistical priors of clean images, such as the Dark Channel Prior (DCP)~\cite{prior1_dcp} and Non-Local Prior (NLP)~\cite{prior2_nlp}, to constrain solutions. However, these priors often fail to generalize across diverse real-world scenarios, resulting in unsatisfactory dehazing outcomes and noticeable artifacts. With the rise of Convolutional Neural Networks (CNNs), numerous learning-based approaches emerged~\cite{dehazenet, pyramid, learningasm3, gdn, gdn+, msbdn, dehamer}. Trained on large-scale datasets, these methods have achieved remarkable advancements. However, collecting large-scale, perfectly aligned real-world hazy and clean image pairs is nearly impossible. Although some real-world datasets have been constructed~\cite{NH-Haze_2020, realdataset1}, their scale and diversity remain insufficient for training robust deep neural networks. Consequently, most current models rely on synthetic datasets, leading to performance degradation when the models are applied to real-world hazy scenes.

Recognizing this limitation, recent studies have increasingly focused on real-world image dehazing. Several studies~\cite{psd, hazeclip} reintroduced prior knowledge to fine-tune pre-trained models, while others~\cite{ridcp, pttd} adapted their inference procedures. Although these methods led to performance gains, they remain significantly dependent on pre-trained models and, consequently, the quality of pre-training data. This dependency highlights the necessity of developing large-scale realistic hazy image generation approaches, which are currently lacking.  Moreover, heavily hazed images involve significant information loss, posing challenges to conventional \textit{enhancement}-based methods that lack generative flexibility to recover such information~\cite{ridcp}. While diffusion models have demonstrated remarkable success in image generation~\cite{stablediffusion, glide, dalle2}, their application to image dehazing remains limited partly due to lengthy sampling processes, underscoring the importance of efficient sampling strategies.

To address these challenges, we propose a  hazing-dehazing pipeline comprising a Realistic Hazy Image Generation framework (HazeGen) and a Diffusion-based Dehazing framework (DiffDehaze). Recent advancements in text-to-image diffusion models~\cite{dalle2, glide, stablediffusion}, trained on extensive datasets containing diverse hazy scenarios, have demonstrated remarkable capabilities in producing high-quality images. By providing suitable prompts, these pre-trained diffusion models can naturally serve as effective generators of realistic hazy images—an insight central to our approach. To enable conditional generation, HazeGen utilizes IRControlNet~\cite{diffbir}, which injects conditional information into the fixed diffusion prior. To further enhance the realism of generated hazy images, HazeGen adopts hybrid conditional and unconditional training objectives and a blended sampling strategy. During sampling, incorporating a small fraction of unconditional predictions can significantly enhance the realism and variety of generated images.

\begin{figure}[t]
	\centering
    \begin{minipage}[h]{0.325\linewidth}
		\centering
		\includegraphics[width=\linewidth]{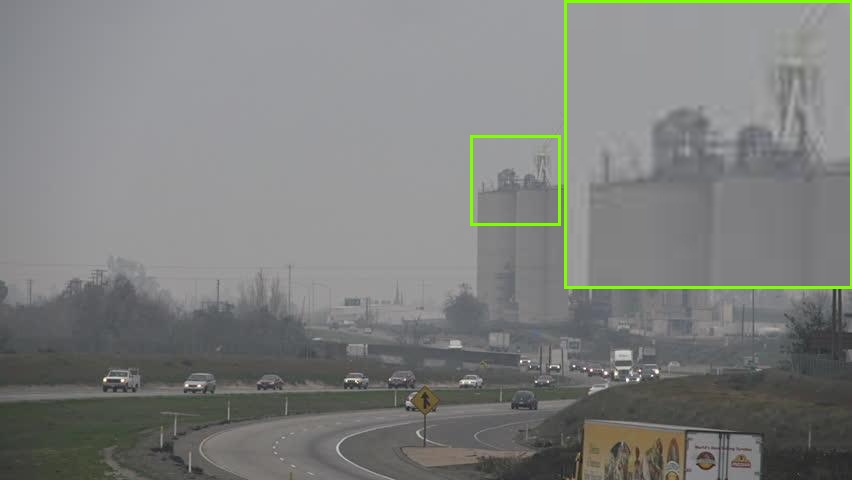}
	\end{minipage}
	\begin{minipage}[h]{0.325\linewidth}
		\centering
		\includegraphics[width=\linewidth]{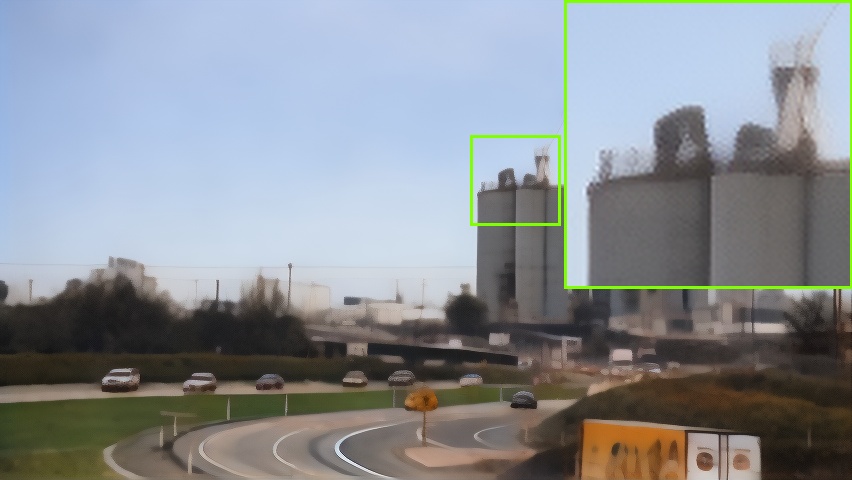}
	\end{minipage}
	\begin{minipage}[h]{0.325\linewidth}
		\centering
		\includegraphics[width=\linewidth]{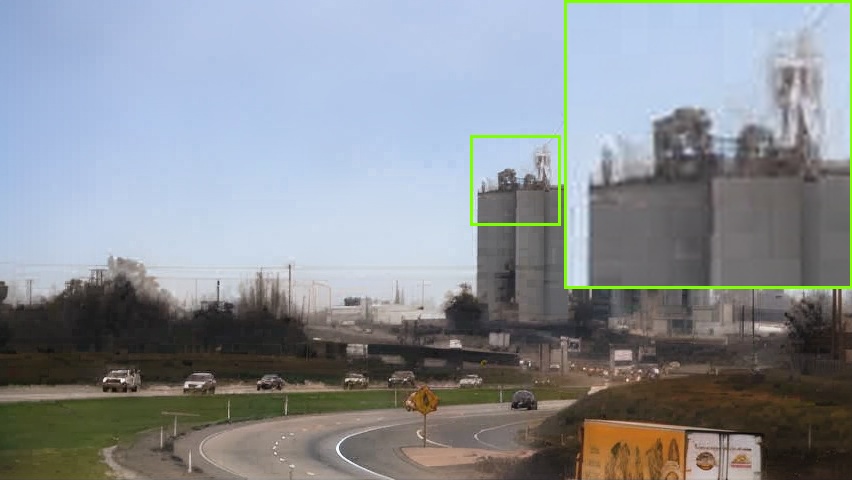}
	\end{minipage}

    \begin{minipage}[h]{0.325\linewidth}
		\centering
		\includegraphics[width=\linewidth]{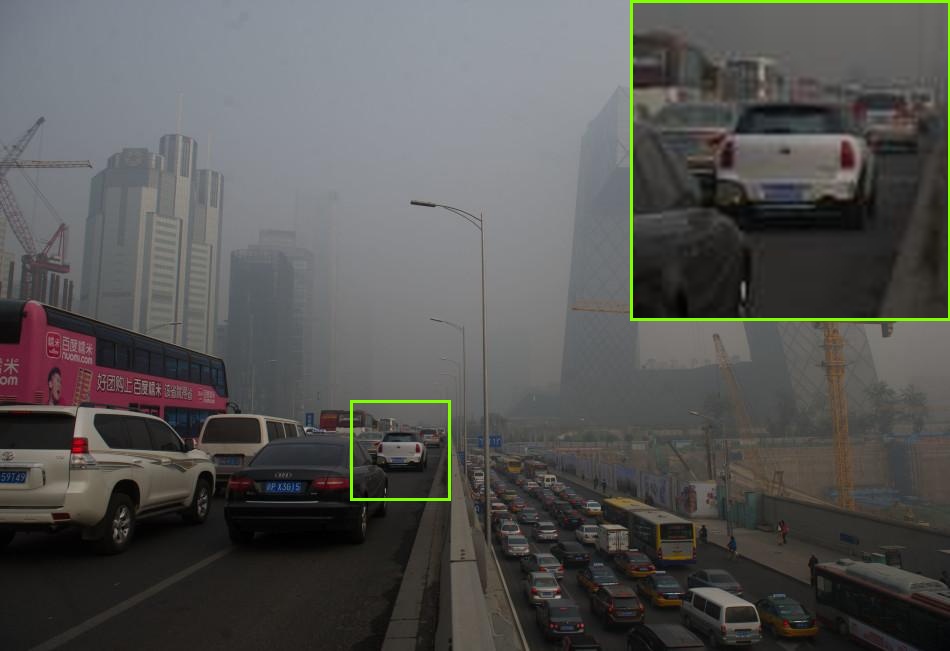}
            \small{(a) Hazy Image}
	\end{minipage}
	\begin{minipage}[h]{0.325\linewidth}
		\centering
		\includegraphics[width=\linewidth]{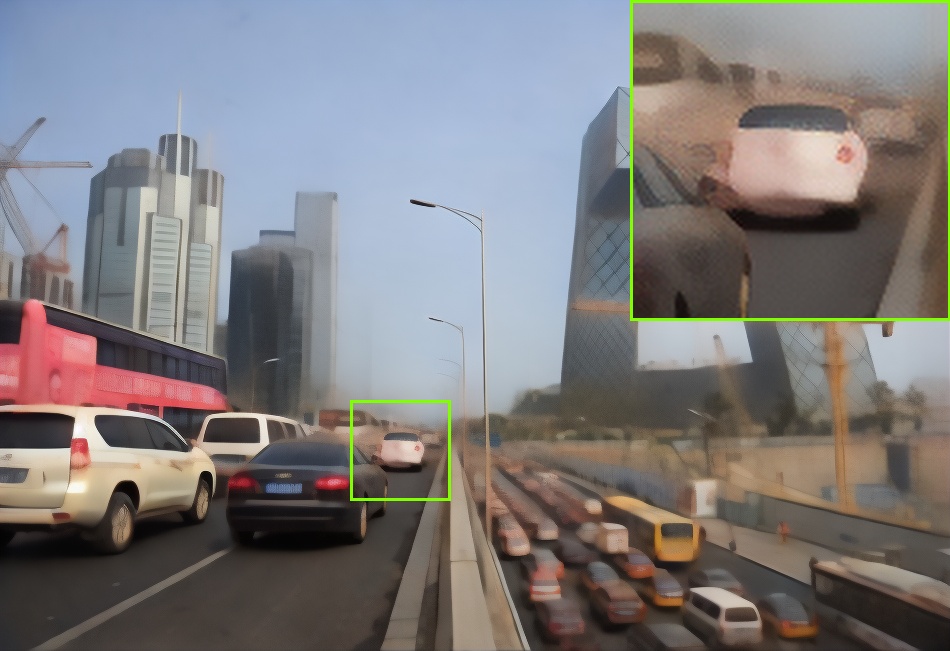}
            \small{(b) Early Prediction}
	\end{minipage}
	\begin{minipage}[h]{0.325\linewidth}
		\centering
		\includegraphics[width=\linewidth]{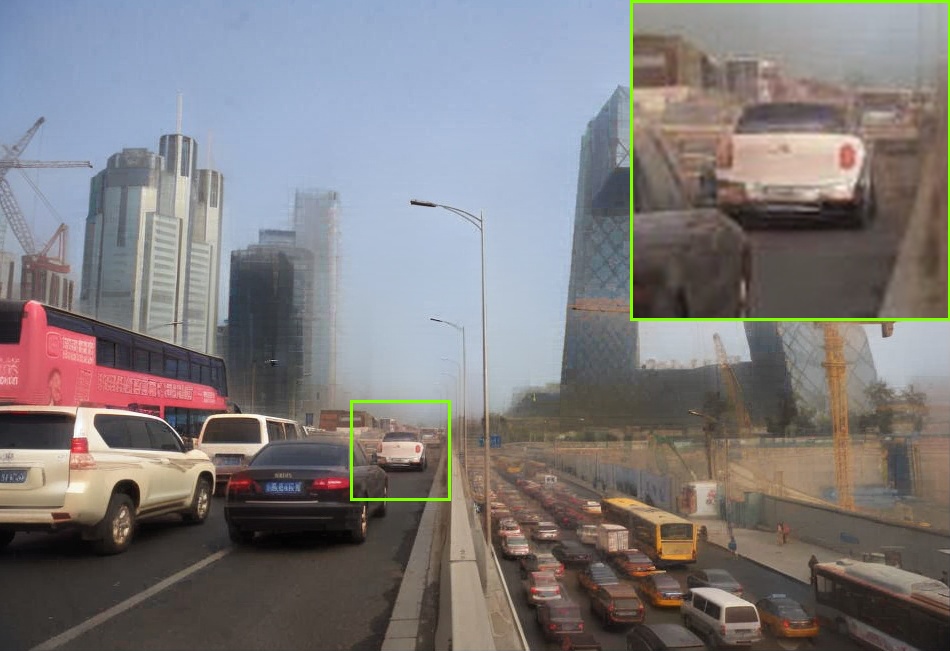}
            \small{(c) Aligned}
	\end{minipage}

	\caption{Visualization of AlignOp’s effect. By aligning the local patch statistics of the hazy image with those of an early diffusion prediction, AlignOp produces a clean and faithful dehazing estimate.}
	\label{fig:AlignOp}
\end{figure}

DiffDehaze is trained using high-quality data produced by HazeGen and employs an Accelerated Fidelity-Preserving Sampling process (AccSamp). The core of AccSamp is the Tiled Statistical Alignment Operation (AlignOp), whose effect is shown in Figure~\ref{fig:AlignOp}. Drawing inspiration from adaptive instance normalization~\cite{adaIN} and image-level normalization~\cite{pttd}, AlignOp substitutes the mean and variance of the hazy image with that of an early diffusion prediction in patches to produce a rough dehazing estimate. Although the early diffusion prediction is blurry, AlignOp can transform it into an effective and faithful dehazing estimate, which allows skipping intermediate sampling steps and advancing directly to a later step. To further enhance fidelity—especially in lightly hazy regions—we additionally equip AccSamp with a haze density-aware fidelity guidance mechanism. 

Our contributions are summarized as follows:

$\diamond$ We introduce a novel Realistic Hazy Image Generation framework (HazeGen) that leverages generative diffusion priors to produce highly realistic hazy training data, significantly boosting the performance of DiffDehaze.

$\diamond$ We propose an Accelerated Fidelity-Preserving Sampling process (AccSamp) for DiffDehaze, which reduces the sampling steps and enhances fidelity of recovered images.

$\diamond$ Extensive quantitative and qualitative experiments validate that our proposed pipeline achieves superior performance compared to state-of-the-art methods.
\section{Related Works}

\subsection{Single Image Dehazing}

\hspace*{\parindent} Early approaches to single-image dehazing~\cite{prior1_dcp, prior2_nlp, prior3_fattal, prior4, prior5_cap, prior6_fattal_clp} typically reconstruct haze-free images by estimating the transmission map in the physical scattering model with statistical priors of clean images. For example, He~\etal~\cite{prior1_dcp} introduced the influential Dark Channel Prior (DCP), based on the observation that the dark channel of a clean image typically approaches zero intensity. Other effective priors include the Non-Local Prior (NLP)~\cite{prior2_nlp} and the Color Attenuation Prior (CAP)~\cite{prior5_cap}. Despite their effectiveness, these handcrafted priors generally struggle to cover the complexity and diversity of real-world imagery, for instance, DCP fails for prominent white objects, resulting in degraded dehazing effect and visible artifacts.

With the emergence of Convolutional Neural Networks (CNNs) and the availability of large-scale synthetic datasets, learning-based dehazing methods~\cite{msbdn, gdn, gdn+, dcmpnet, dehamer, mbtaylorformer, ffanet, dehazedct} have become increasingly popular, paralleling advances in other low-level vision tasks~\cite{han2025glare, dong2024ecmamba, zhou2024lowlightimageenhancementgenerative, Jiang_2024_ECCV, FastLLVE, Situation-adaptive, fu2024attentionlutattentionfusionbasedcanonical}. Particularly, Liu~\etal~\cite{gdn} proposed an attention-based multi-scale grid network, and Qiu~\etal~\cite{mbtaylorformer} developed a Transformer variant with linear computational complexity to efficiently harness Transformer architectures for image dehazing. However, although learning-based techniques achieve superior performance on synthetic datasets, their reliance on synthetic training data and the substantial domain gap between synthetic and real-world hazy images cause significant performance degradation on real-world hazy images.

\subsection{Real-World Image Dehazing}
\hspace*{\parindent}Due to the challenges and practical significance of real-world image dehazing, recent research has increasingly emphasized methods tailored for real-world scenarios. Early studies~\cite{d4, dad} explored CycleGAN-based frameworks. Shao~\etal~\cite{dad}, for example, introduced a CycleGAN-based domain adaptation approach to map images from the synthetic domain to the real domain. Nevertheless, these frameworks usually have complex and unstable training processes, as well as mode collapse issues. Another popular research direction~\cite{semisupervised, psd} is to integrate prior knowledge into the fine-tuning process of pre-trained models. For instance, Chen~\etal~\cite{psd} proposed PSD, a framework converting pre-trained dehazing networks into physically informed ones and fine-tuning them via three statistical priors. Other studies~\cite{ridcp, pttd} target improvements during inference. For example, Chen~\etal~\cite{pttd} proposed a feature adaptation module designed to recalibrate encoder features during inference. Despite the advances, these techniques remain inherently dependent on the quality of the pre-training data, underscoring the necessity of robust methods capable of generating realistic and diverse hazy images. Prior attempts~\cite{reside, ridcp} have utilized the physical scattering model and depth maps for haze synthesis. However, the physical scattering model can't adequately represent the intricate real-world haze formation process, resulting in unrealistic and homogeneous hazy images. In contrast to explicit physical modeling, our work explores robust generative priors embedded in a pre-trained text-to-image diffusion model.

\section{Methodology}

\hspace*{\parindent}The central idea of this work is to leverage the generative diffusion prior of natural hazy images. HazeGen and DiffDehaze are built upon the architectural framework introduced by DiffBIR~\cite{diffbir}, employing a fixed Stable Diffusion model and a IRControlNet for conditional information injection. IRControlNet is initialized from the UNet's encoder and modulates features in its decoder. HazeGen further applies specialized hybrid training and blended sampling strategies, whereas DiffDehaze adopts the AccSamp sampling process for enhanced efficiency and fidelity. In the following sections, we briefly review diffusion models and then present the detailed methodologies of HazeGen and DiffDehaze.

\subsection{Preliminary: Diffusion Models}
\hspace*{\parindent} Since diffusion models form the foundation for HazeGen and DiffDehaze, we briefly review key concepts of conditional denoising diffusion models~\cite{ddpm, sr3}. HazeGen aims to model the conditional distribution $P(x|y)$, where $x$ and $y$ denote corresponding hazy and clean images, while DiffDehaze inversely models $P(y|x)$. Diffusion models have two fundamental processes: the forward diffusion process and the reverse denoising process.

{\noindent\textbf{The Forward Process.}}
The forward process progressively adds Gaussian noise to an encoded target image $z_0 = z$ according to a predefined variance schedule ${\beta_t}$. This process can be succinctly formulated as:
\begin{equation}
  z_t = \sqrt{\bar{\alpha}_t}z_0+\sqrt{1-\bar{\alpha}_t}\epsilon,
  \label{eq:forward}
\end{equation}
where $z_t$ is the noised image, $\alpha_t=1-\beta_t$, $\bar{\alpha}_t=\sum^t_{s=1}\alpha_s$ and $\epsilon\sim\mathcal{N}(0, \mathbf{1})$.

\noindent\textbf{The Reverse Process.}
The reverse process aims to reconstruct the original image by denoising. Specifically, given encoded condition $c$, a model $\epsilon_\theta$ learns to predict the added noise $\epsilon$ at each timestep $t$. At a randomly sampled timestep $t$, noise $\epsilon$ is introduced to produce $z_t$. The model is trained to minimize the simplified denoising objective~\cite{ddpm}:
\begin{equation}
  \mathcal{L}=\mathbb{E}_{z,c,t,\epsilon\sim\mathcal{N}(0, \mathbf{1})}[||\epsilon-\epsilon_\theta(z_t,c,t)||_2^2].
  \label{eq:simplifed_obj}
\end{equation}

To leverage rich generative priors, both HazeGen and DiffDehaze are built upon the pre-trained Stable Diffusion model~\cite{stablediffusion}. Unlike traditional diffusion methods operating in RGB pixel space, Stable Diffusion performs the diffusion and denoising processes in a low-dimensional latent space created by a separate autoencoder.

      
      

\subsection{Realistic Hazy Image Generation Framework}
\hspace*{\parindent}Though equipped with powerful generative priors, HazeGen must effectively incorporate conditional information from clean images. A straightforward solution would be training HazeGen on paired synthetic images. However, we observe that this method rapidly degenerates the generative priors, as the simple haze synthesis with the physical scattering model is easily recognizable and thus is learned by the model. As a result, generated hazy images closely resemble synthetic data, causing a significant domain gap with real-world images. To alleviate this issue, we propose specialized hybrid training and blended sampling algorithms.

\noindent\textbf{Hybrid Training.}
To preserve the ability to generate realistic hazy images while gradually enhancing content consistency, we introduce the hybrid training objective. Specifically, the conditional generation objective uses synthetic image pairs, guiding the model to build content relationships between generated hazy images and corresponding clean images. Conversely, the unconditional objective employs unlabeled real-world hazy images, helping HazeGen maintain and further enhance its capability for realistic haze generation, thus preventing catastrophic forgetting. Combining the two objectives, the hybrid training loss is:
\begin{align}
    \mathcal{L} = &\,  p\ \mathbb{E}_{z^s,c^s,t,\epsilon} 
    \left[ \|\epsilon - \epsilon_\theta(z^s_t, c^s, t) \|_2^2 \right] \nonumber \\
    &+ (1 - p) \ \mathbb{E}_{z^r,t,\epsilon} 
    \left[ \|\epsilon - \epsilon_\theta(z^r_t, \varnothing, t) \|_2^2 \right],
    \label{eq:simplified_obj}
\end{align}
where encoded synthetic image pairs are represented by $(z^s, c^s)$, while encoded real-world hazy images are denoted as $z^r$. $p$ is a tradeoff parameter determining the probability to apply the conditional objective. 


\begin{algorithm}[tb]
  \caption{Blended sampling algorithm, given the denoising model $\bepsilon_\theta$ and the VAE's encoder $\mathcal{E}$ and decoder $\mathcal{D}$ } \label{alg:sampling}
  \begin{algorithmic}[1]
    \Require $w$: mixture coefficient, $y$: a clean image
    \State $\bz_{T} \sim \mathcal{N}(\bzero, \bI)$
    \For{$t=T, \dotsc, 1$}

        $\!\!\triangleright$ Blended noise prediction
      \State $\hat{\bepsilon} = w\;\bepsilon_\theta(\bz_{t}, \mathcal{E}(y), t) + (1-w)\;\bepsilon_{\theta}(\bz_{t}, \varnothing, t)$ 
      
      $\!\!\triangleright$ Sampling step
      \State $\bepsilon \sim \mathcal{N}(\bzero, \bI)$ 
      \State $\bz_{t-1} \gets \frac{1}{\sqrt{\alpha_t}} \left( \bz_t-\frac{1-\alpha_t}{\sqrt{1-\bar{\alpha}_t}}\hat{\bepsilon} \right) + \frac{1-\bar{\alpha}_{t-1}}{1-\bar{\alpha}_t}(1-\alpha_t) \bepsilon$       
    \EndFor
    \State \textbf{return} $\mathcal{D}(\bz_{0})$
  \end{algorithmic}
\end{algorithm}

\begin{figure}[t]
	\centering
    \begin{minipage}[h]{0.24\linewidth}
		\centering
		\includegraphics[width=\linewidth]{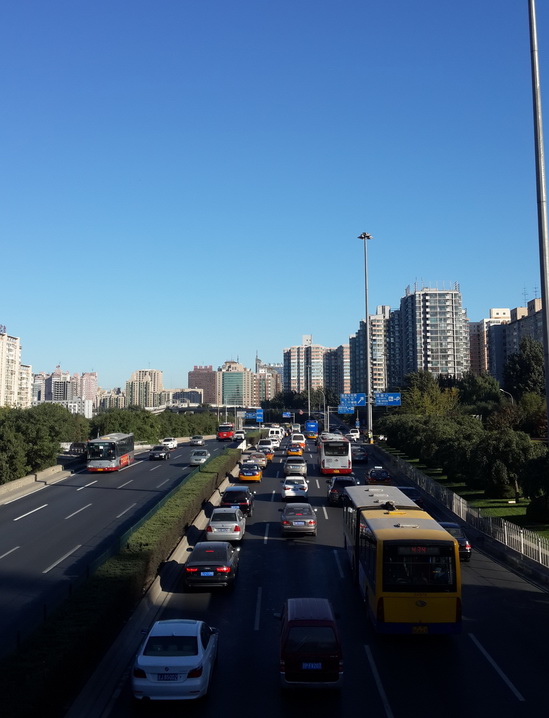}
	\end{minipage}
	\begin{minipage}[h]{0.24\linewidth}
		\centering
		\includegraphics[width=\linewidth]{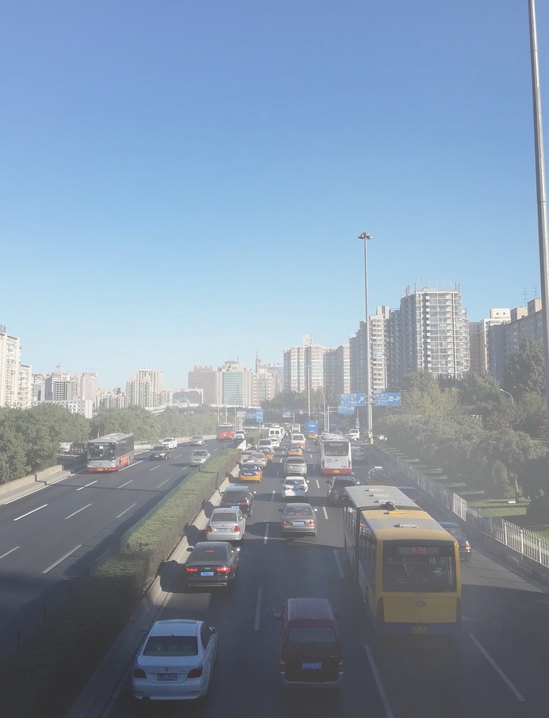}
	\end{minipage}
	\begin{minipage}[h]{0.24\linewidth}
		\centering
		\includegraphics[width=\linewidth]{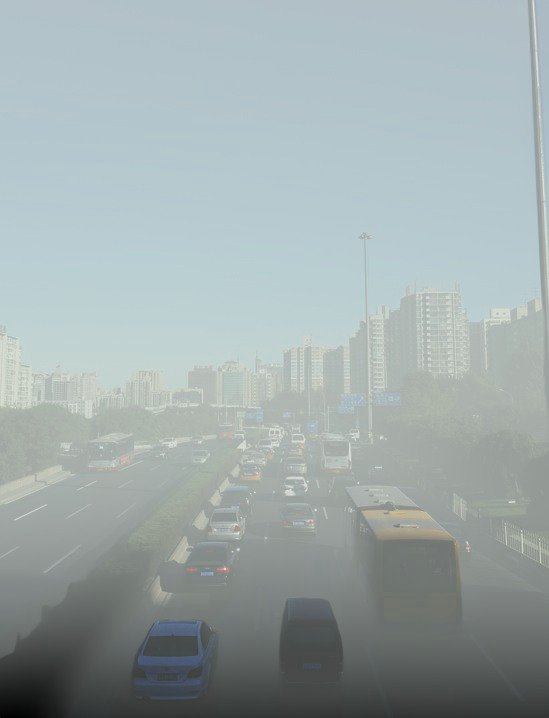}
	\end{minipage}
    	\begin{minipage}[h]{0.24\linewidth}
		\centering
		\includegraphics[width=\linewidth]{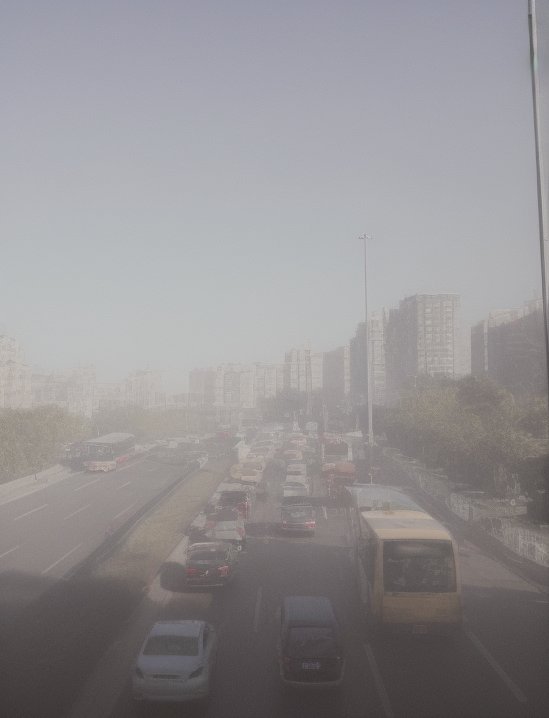}
	\end{minipage}

    \begin{minipage}[h]{0.24\linewidth}
		\centering
		\includegraphics[width=\linewidth]{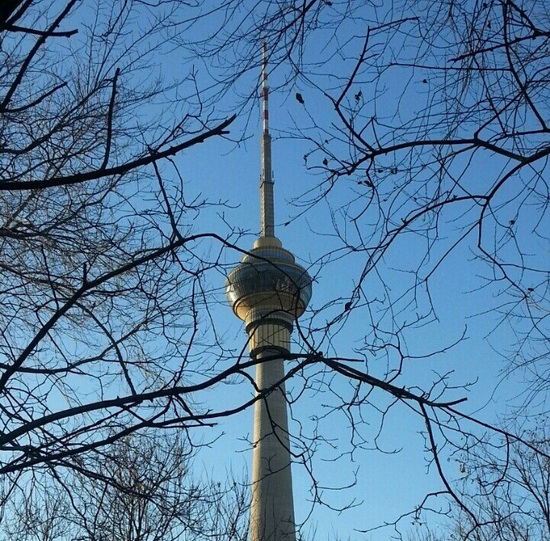}
	\end{minipage}
	\begin{minipage}[h]{0.24\linewidth}
		\centering
		\includegraphics[width=\linewidth]{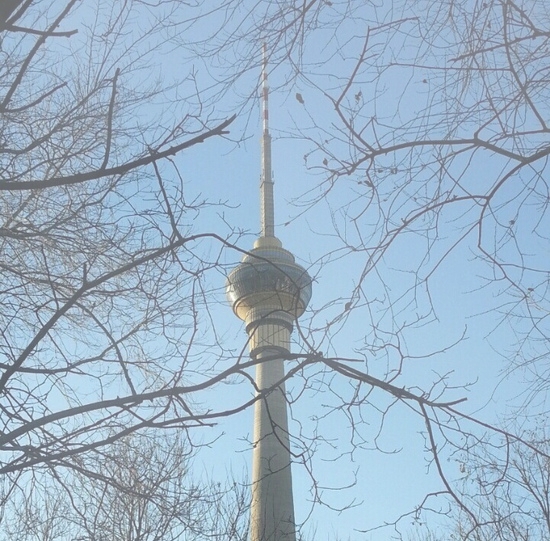}
	\end{minipage}
	\begin{minipage}[h]{0.24\linewidth}
		\centering
		\includegraphics[width=\linewidth]{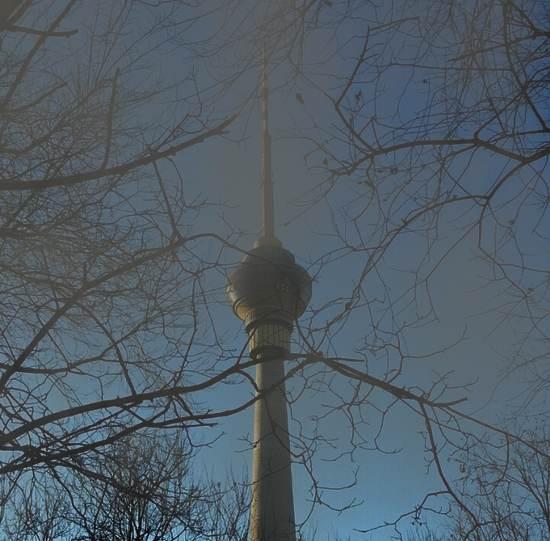}
	\end{minipage}
    	\begin{minipage}[h]{0.24\linewidth}
		\centering
		\includegraphics[width=\linewidth]{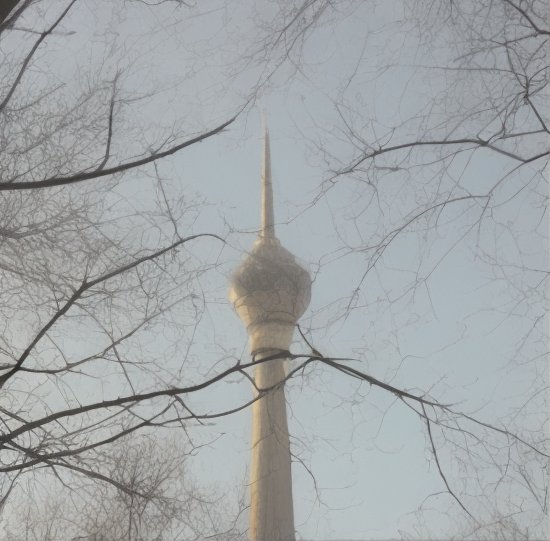}
	\end{minipage}
    
    \begin{minipage}[h]{0.24\linewidth}
		\centering
		\includegraphics[width=\linewidth]{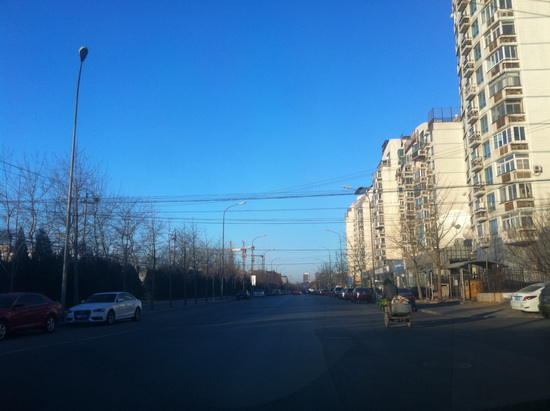}
            \small{(a) Input}
	\end{minipage}
    	\begin{minipage}[h]{0.24\linewidth}
		\centering
		\includegraphics[width=\linewidth]{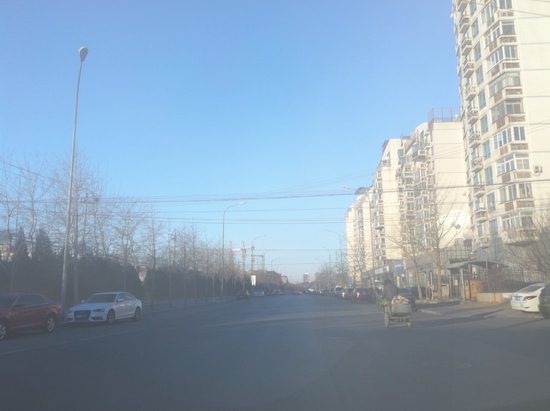}
            \small{(b) OTS~\cite{reside}}
	\end{minipage}
	\begin{minipage}[h]{0.24\linewidth}
		\centering
		\includegraphics[width=\linewidth]{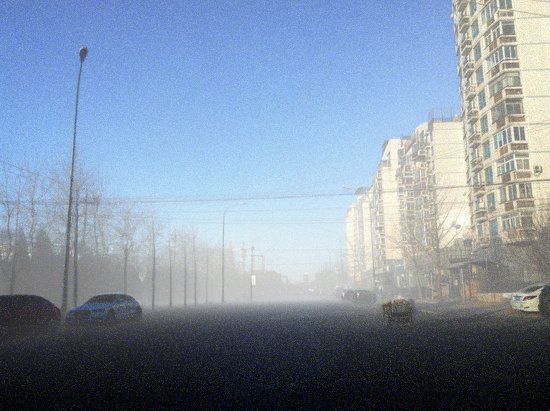}
            \small{(c) RIDCP~\cite{ridcp}}
	\end{minipage}
	\begin{minipage}[h]{0.24\linewidth}
		\centering
		\includegraphics[width=\linewidth]{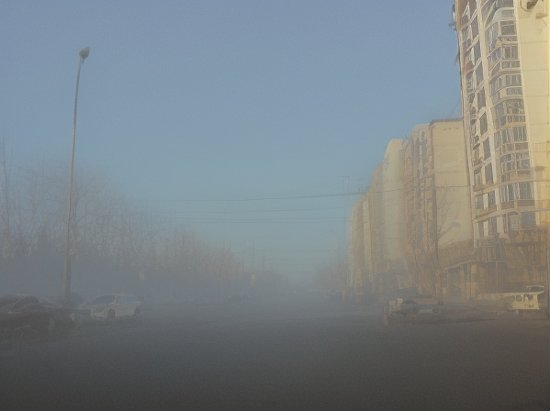}
            \small{(d) Ours}
	\end{minipage}
	\caption{Visual comparisons between hazy images generated by HazeGen and synthetic images from OTS~\cite{reside} and the phenomenological degradation pipeline of RIDCP~\cite{ridcp}.}
	\label{fig:comparison_hazy}
\end{figure}

\noindent\textbf{Blended Sampling.}
Although hybrid training significantly mitigates overfitting to synthetic data, we further propose a blended sampling strategy. This strategy incorporates a small fraction of unconditional predictions into the conditional noise predictions at each sampling step. The advantages of blended sampling are two-fold: (1) unconditional predictions, leveraging intact generative capabilities, can effectively compensate for the learned defects in conditional predictions; and (2) the overall diversity of generated images is notably improved. The detailed blended sampling procedure is provided in Algorithm~\ref{alg:sampling}.

An analogy can be made between our approach and the classifier-free diffusion guidance~\cite{classifierfree}. However, distinct from classifier-free guidance, our primary motivation is to mitigate the adverse impact of low-quality training data and to enhance sampling diversity. 

Visual comparisons between hazy images generated by HazeGen and two physical scattering model-based synthesis, OTS~\cite{reside} and the RIDCP degradation pipeline~\cite{ridcp}, are presented in Figure~\ref{fig:comparison_hazy}. The hazy images from OTS exhibit only mild haze, while those from the RIDCP degradation pipeline suffer from unrealistic abrupt haze-density changes around nearby objects and artifacts caused by inaccurate depth estimation. In contrast, HazeGen generates hazy images with more realistic and visually consistent haze.
\begin{figure*}[t]
  \centering
  \includegraphics[width=0.98\linewidth]{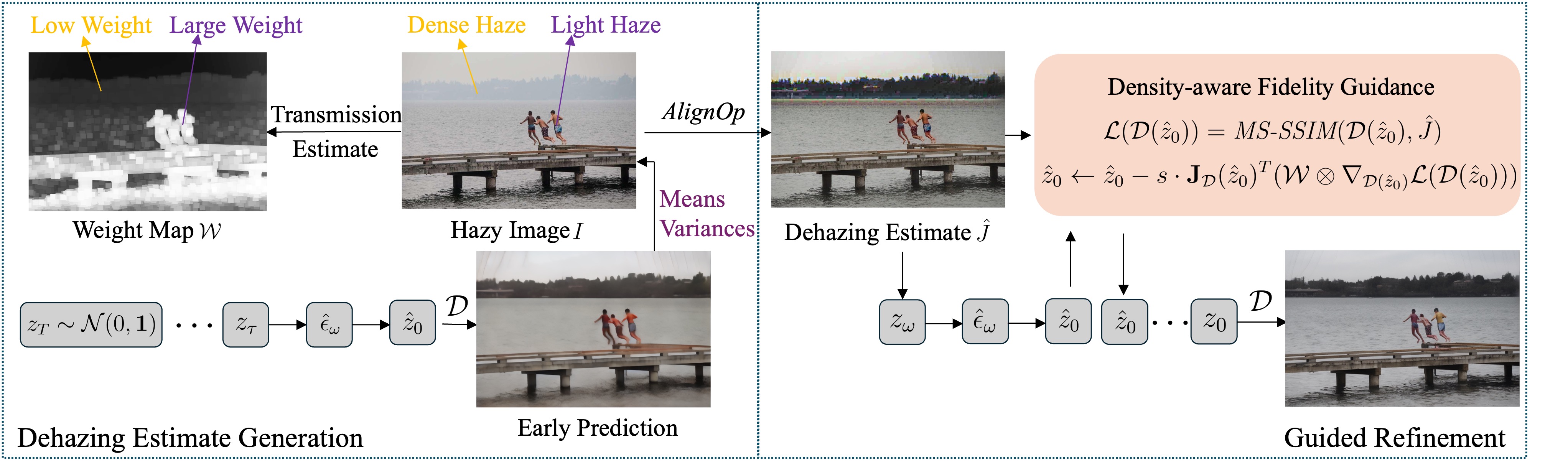}
  \caption{Overview of the AccSamp sampling process. The accelerated sampling process consists of two stages: the dehazing estimate generation stage and the guided refinement stage. In the initial stage (timesteps $T$ to $\tau$), AlignOp transforms a blurry early diffusion prediction into a detailed and faithful dehazing estimate. In the subsequent refinement stage (the final $\omega$ steps), additional vivid details are generated under density-aware fidelity guidance. Intermediate sampling steps between $\tau$ and $\omega$ are skipped to enhance efficiency.
}
  \label{fig:method}
\end{figure*}

\subsection{Diffusion-based Dehazing Framework}
\hspace*{\parindent} DiffDehaze is trained with high-quality data generated by HazeGen and can thus produce high-quality dehazing results. To reduce the computational cost of lengthy sampling processes and improve fidelity, we propose the Accelerated Fidelity-Preserving Sampling process, AccSamp. As illustrated in Figure~\ref{fig:method}, the sampling process is divided into two stages: the dehazing estimate generation stage and the guided refinement stage. Specifically, the dehazing estimate generation stage covers initial sampling steps from $T$ down to $\tau$, and the guided refinement stage encompasses the final $\omega$ steps. The intermediate timesteps between $\tau$ and $\omega$ can be skipped to enhance sampling efficiency.

\noindent\textbf{Dehazing Estimate Generation Stage.}
The core of AccSamp lies in the fast dehazing estimate generation enabled by AlignOp. Inspired by style transfer~\cite{adaIN} and image-level normalization~\cite{pttd}, we observe that aligning local mean and variance between hazy and corresponding clean image patches yields a reliable preliminary dehazing estimate. This insight aligns well with the physical scattering model: within local regions of approximately uniform scene depth—implying a nearly constant transmission map in Equation~\ref{eq:asm}—the hazy image patch essentially represents a scaled and shifted version of the clean patch. Consequently, statistical alignment approximates a reversal of the haze formation process. Importantly, as only the statistics of clean image patches are required, their details are unnecessary. Therefore, we equivalently utilize an early-stage diffusion prediction, based by the observation that conditional diffusion models establish the overall color distribution within a small fraction of diffusion steps. Through AlignOp, a blurry, low-quality early-stage prediction at timestep $\tau$ can be transformed to a satisfactory coarse dehazing estimate with effective dehazing performance while preserving details from the hazy image. 

Specifically, the predicted mean at timestep $t=\tau$, denoted as  $\hat{z}_0^{(\tau)}$, can be computed by reversing Equation~\ref{eq:forward}:
\begin{equation}
  \hat{z}_0^{(\tau)} = \frac{z_\tau-\sqrt{1-\bar{\alpha}_\tau}\epsilon_\theta(z_\tau, c, \tau)}{\sqrt{\bar{\alpha}_\tau}},
  \label{eq:predict_mean}
\end{equation}
where $c=\mathcal{E}(x)$ represents the encoded hazy image.

Overlapping patches from the hazy image $x$ and the early prediction $\mathcal{D}(\hat{z}_0^{(\tau)})$ are extracted using a sliding window with patch size $k \times k$ and stride $d$. These patches are denoted as $\{p_i^x\}$ and $\{p_i^r\}$, respectively. For each pair of patches $(p_i^x, p_i^r)$, the aligned patch $p_i^{\hat{y}}$ is computed as
\begin{equation}
  p_i^{\hat{y}}\:=\:\frac{p_i^x-\mu(p_i^x)}{\sigma(p_i^x)}\cdot\sigma(p_i^r)+\mu(p_i^r),
  \label{eq:psao}
\end{equation}
where $\mu(\cdot)$ and $\sigma(\cdot)$ compute the channel-wise mean and standard deviation. The dehazing estimate $\hat{y}$ is then obtained by assembling these aligned patches and averaging their overlapping regions.

Since the dehazing estimate retains details from the hazy image, its difference from the actual underlying clean image is relatively minor. Consequently, we assume that after adding Gaussian noise at timestep $\omega$, the distribution of the dehazing estimate closely approximates that of the true clean image. Thus, $z_\omega$ can be directly approximated by adding noise to $\mathcal{E}(\hat{y})$ with $\epsilon \sim \mathcal{N}(0, \mathbf{1})$:
\begin{equation}
  z_\omega = \sqrt{\bar{\alpha}_\omega}\;\mathcal{E}(\hat{y})+\sqrt{1-\bar{\alpha}_\omega}\epsilon.
  \label{eq:forward_omega}
\end{equation}

\begin{figure*}[t]
	\centering
    \begin{minipage}[h]{0.138\linewidth}
		\centering
		\includegraphics[width=\linewidth]{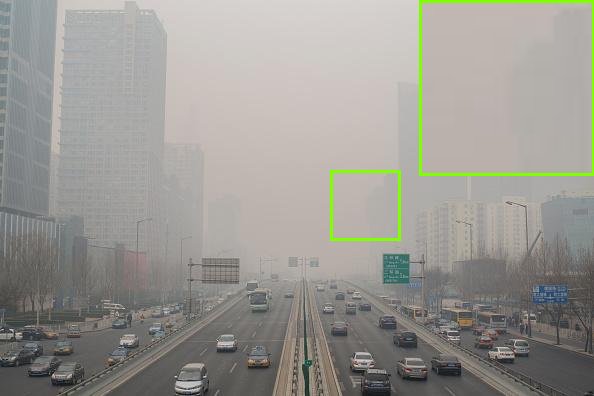}
	\end{minipage}
	\begin{minipage}[h]{0.138\linewidth}
		\centering
		\includegraphics[width=\linewidth]{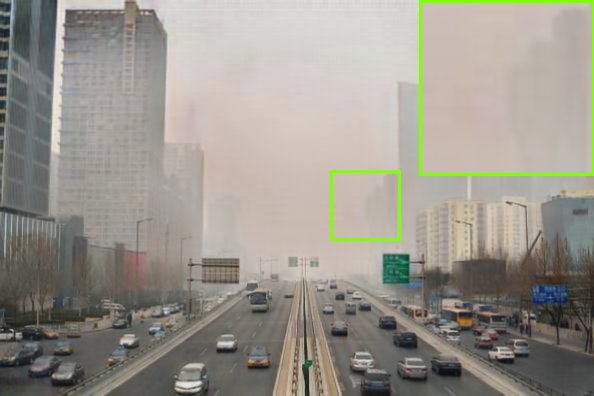}
	\end{minipage}
	\begin{minipage}[h]{0.138\linewidth}
		\centering
		\includegraphics[width=\linewidth]{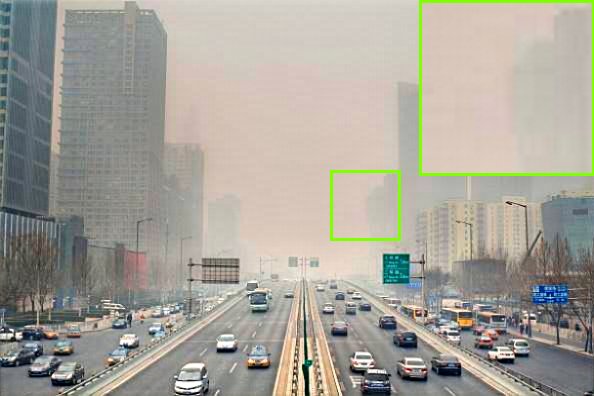}
	\end{minipage}
	\begin{minipage}[h]{0.138\linewidth}
		\centering
		\includegraphics[width=\linewidth]{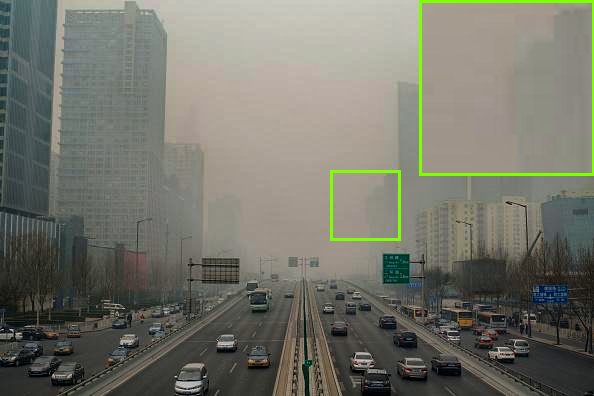}
	\end{minipage}
	\begin{minipage}[h]{0.138\linewidth}
		\centering
		\includegraphics[width=\linewidth]{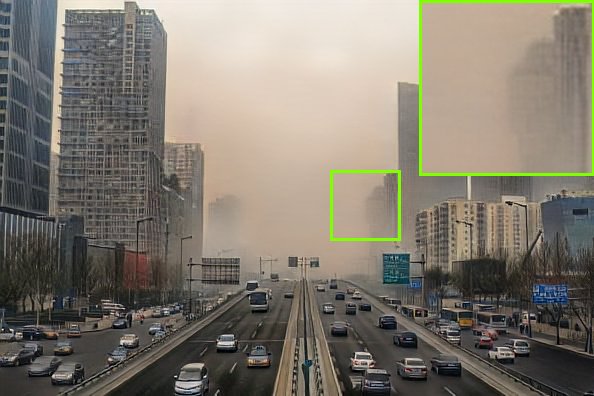}
	\end{minipage}
	\begin{minipage}[h]{0.138\linewidth}
		\centering
		\includegraphics[width=\linewidth]{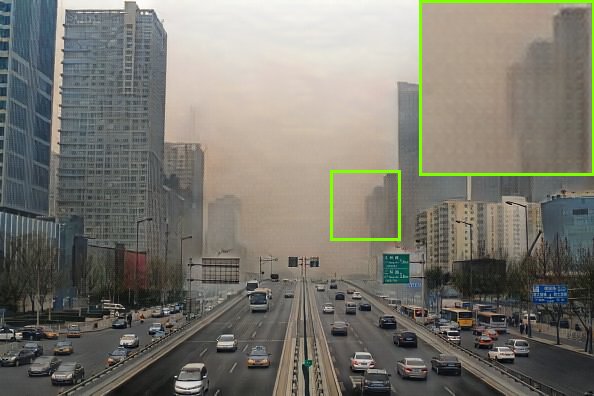}
	\end{minipage}
	\begin{minipage}[h]{0.138\linewidth}
		\centering
		\includegraphics[width=\linewidth]{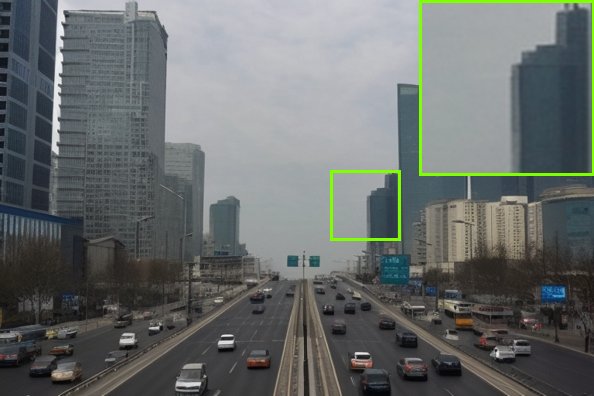}
	\end{minipage}
 \begin{minipage}[h]{0.138\linewidth}
		\centering
		\includegraphics[width=\linewidth]{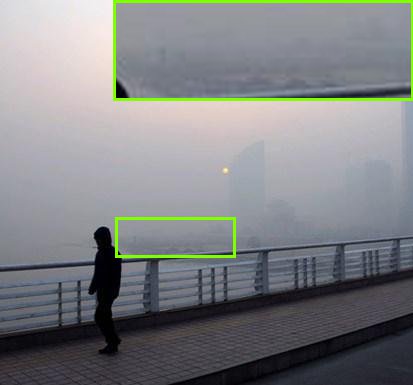}
	\end{minipage}
	\begin{minipage}[h]{0.138\linewidth}
		\centering
		\includegraphics[width=\linewidth]{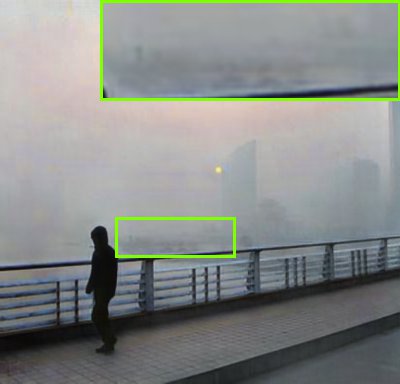}
	\end{minipage}
	\begin{minipage}[h]{0.138\linewidth}
		\centering
		\includegraphics[width=\linewidth]{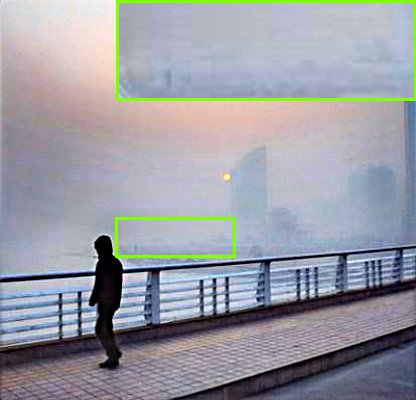}
	\end{minipage}
	\begin{minipage}[h]{0.138\linewidth}
		\centering
		\includegraphics[width=\linewidth]{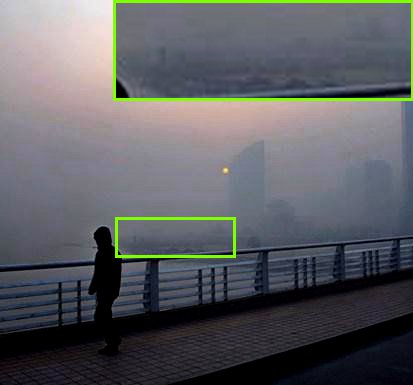}
	\end{minipage}
	\begin{minipage}[h]{0.138\linewidth}
		\centering
		\includegraphics[width=\linewidth]{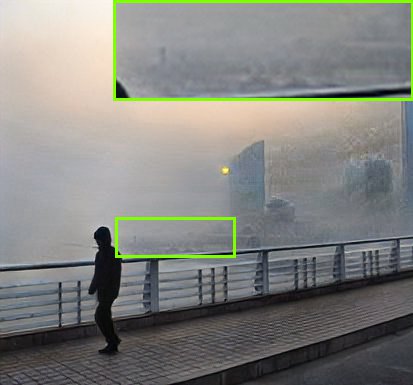}
	\end{minipage}
	\begin{minipage}[h]{0.138\linewidth}
		\centering
		\includegraphics[width=\linewidth]{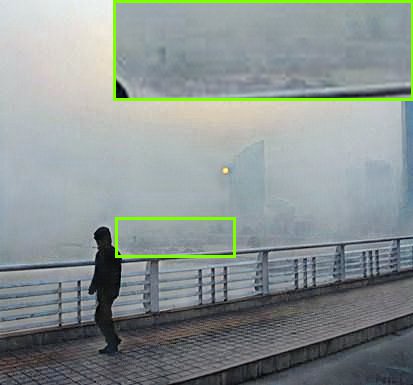}
	\end{minipage}
	\begin{minipage}[h]{0.138\linewidth}
		\centering
		\includegraphics[width=\linewidth]{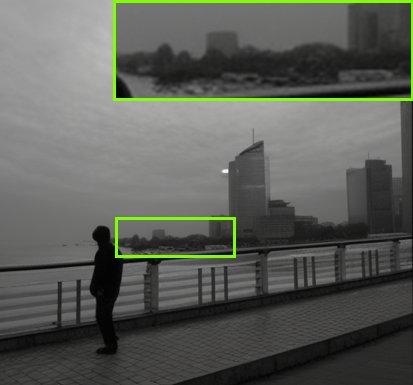}
	\end{minipage}
 \begin{minipage}[h]{0.138\linewidth}
		\centering
		\includegraphics[width=\linewidth]{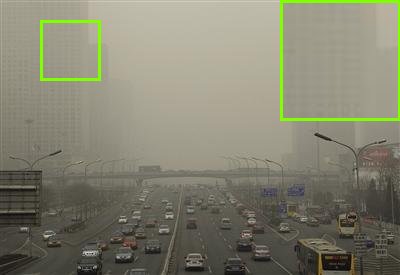}
	\end{minipage}
	\begin{minipage}[h]{0.138\linewidth}
		\centering
		\includegraphics[width=\linewidth]{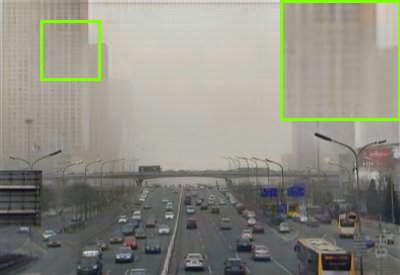}
	\end{minipage}
	\begin{minipage}[h]{0.138\linewidth}
		\centering
		\includegraphics[width=\linewidth]{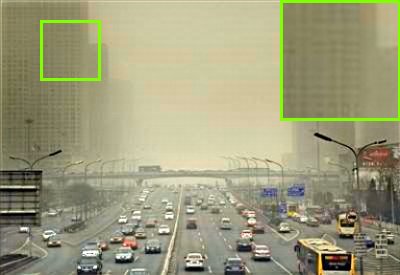}
	\end{minipage}
	\begin{minipage}[h]{0.138\linewidth}
		\centering
		\includegraphics[width=\linewidth]{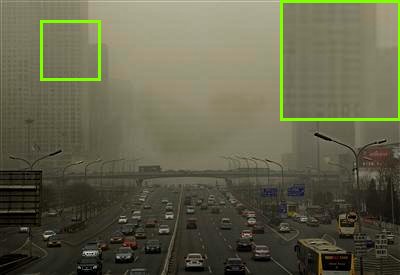}
	\end{minipage}
	\begin{minipage}[h]{0.138\linewidth}
		\centering
		\includegraphics[width=\linewidth]{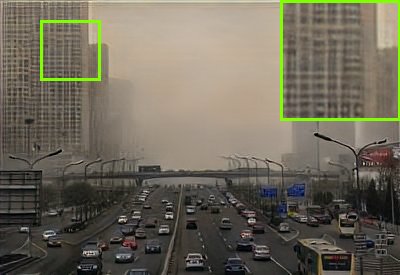}
	\end{minipage}
	\begin{minipage}[h]{0.138\linewidth}
		\centering
		\includegraphics[width=\linewidth]{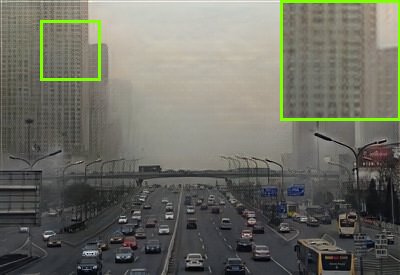}
	\end{minipage}
	\begin{minipage}[h]{0.138\linewidth}
		\centering
		\includegraphics[width=\linewidth]{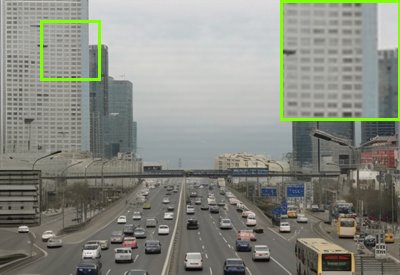}
	\end{minipage}
 \begin{minipage}[h]{0.138\linewidth}
		\centering
		\includegraphics[width=\linewidth]{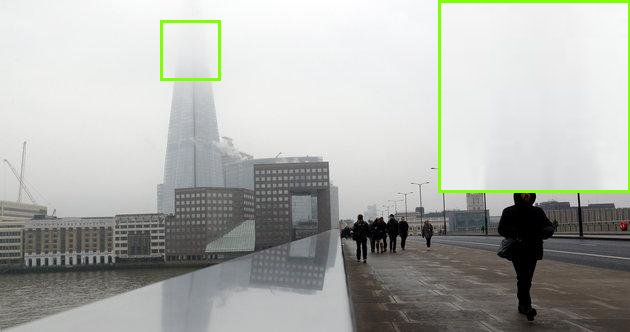}
	\end{minipage}
	\begin{minipage}[h]{0.138\linewidth}
		\centering
		\includegraphics[width=\linewidth]{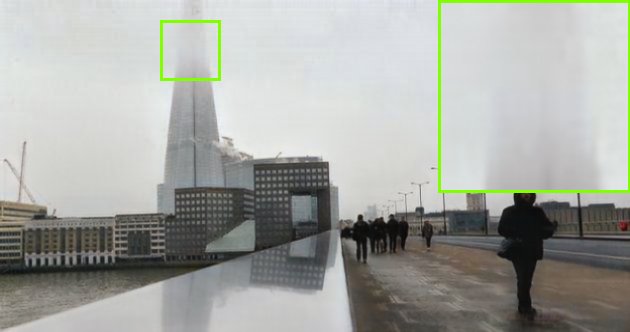}
	\end{minipage}
	\begin{minipage}[h]{0.138\linewidth}
		\centering
		\includegraphics[width=\linewidth]{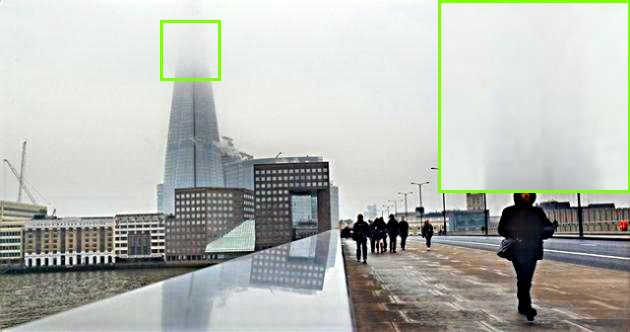}
	\end{minipage}
	\begin{minipage}[h]{0.138\linewidth}
		\centering
		\includegraphics[width=\linewidth]{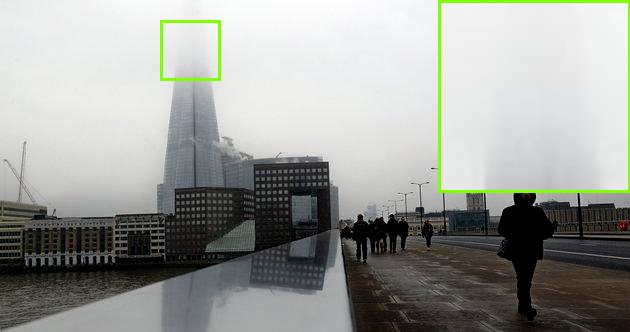}
	\end{minipage}
	\begin{minipage}[h]{0.138\linewidth}
		\centering
		\includegraphics[width=\linewidth]{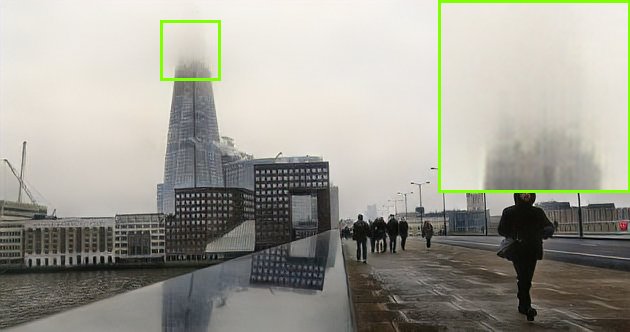}
	\end{minipage}
	\begin{minipage}[h]{0.138\linewidth}
		\centering
		\includegraphics[width=\linewidth]{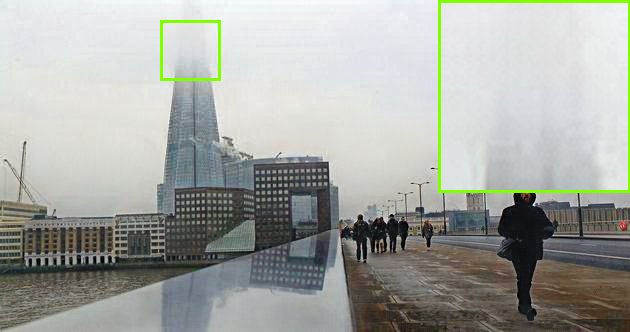}
	\end{minipage}
	\begin{minipage}[h]{0.138\linewidth}
		\centering
		\includegraphics[width=\linewidth]{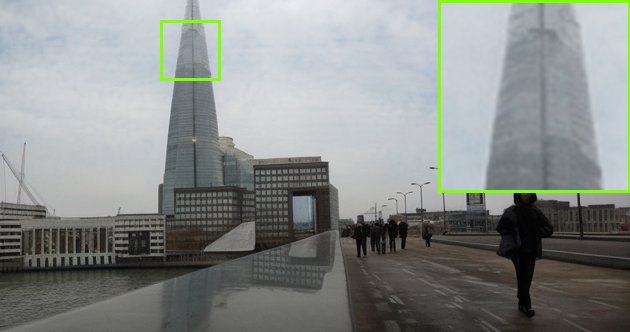}
	\end{minipage}
    \begin{minipage}[h]{0.138\linewidth}
		\centering
		\includegraphics[width=\linewidth]{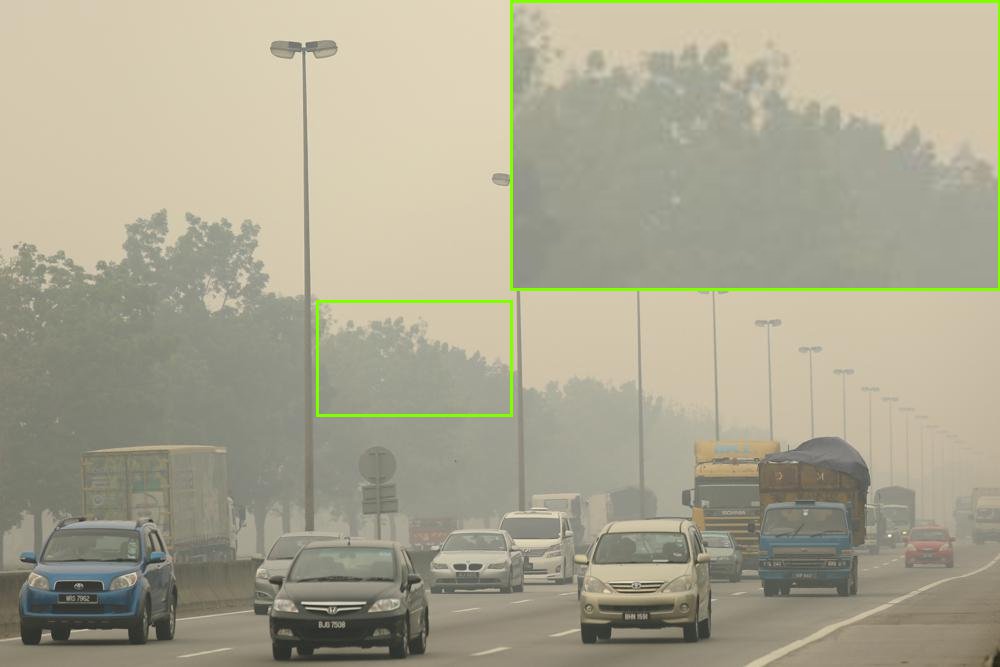}
            \small{(a) Input}
	\end{minipage}
	\begin{minipage}[h]{0.138\linewidth}
		\centering
		\includegraphics[width=\linewidth]{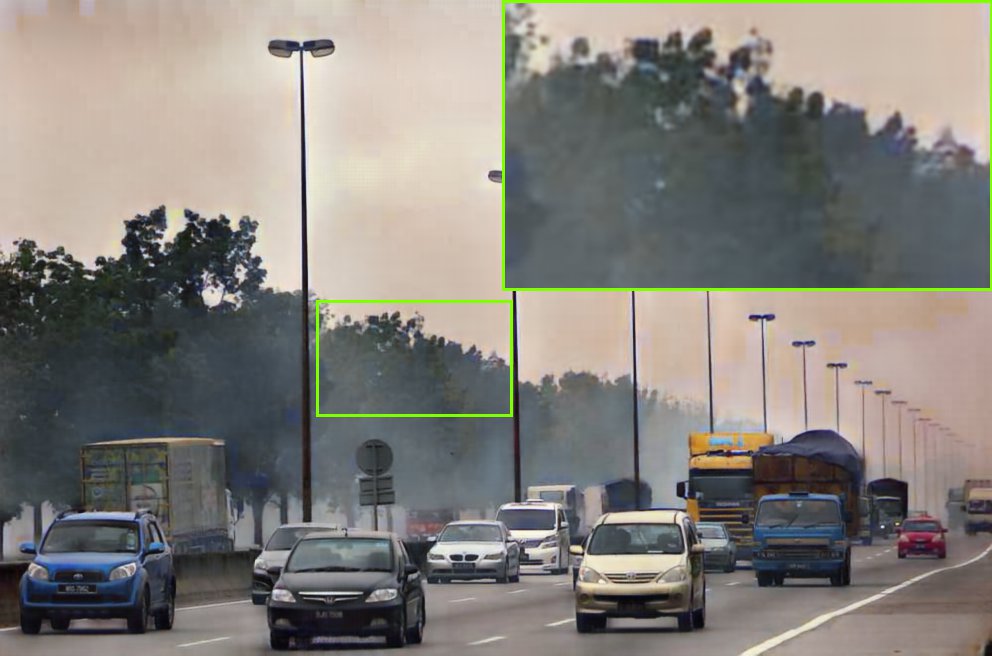}
            \small{(b) DAD~\cite{dad}}
	\end{minipage}
	\begin{minipage}[h]{0.138\linewidth}
		\centering
		\includegraphics[width=\linewidth]{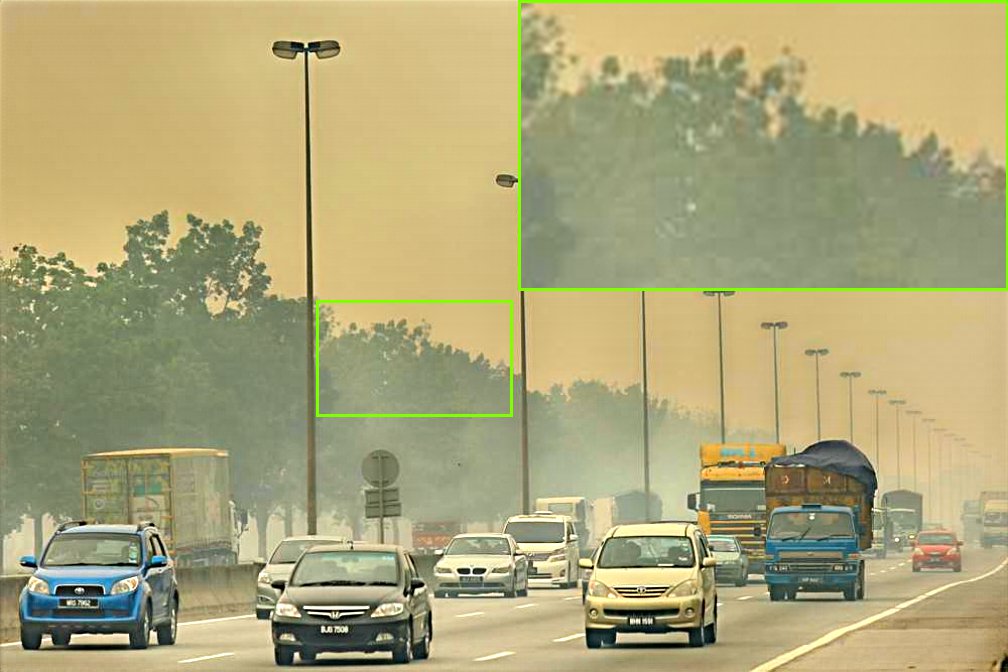}
            \small{(c) PSD~\cite{psd}}
	\end{minipage}
	\begin{minipage}[h]{0.138\linewidth}
		\centering
		\includegraphics[width=\linewidth]{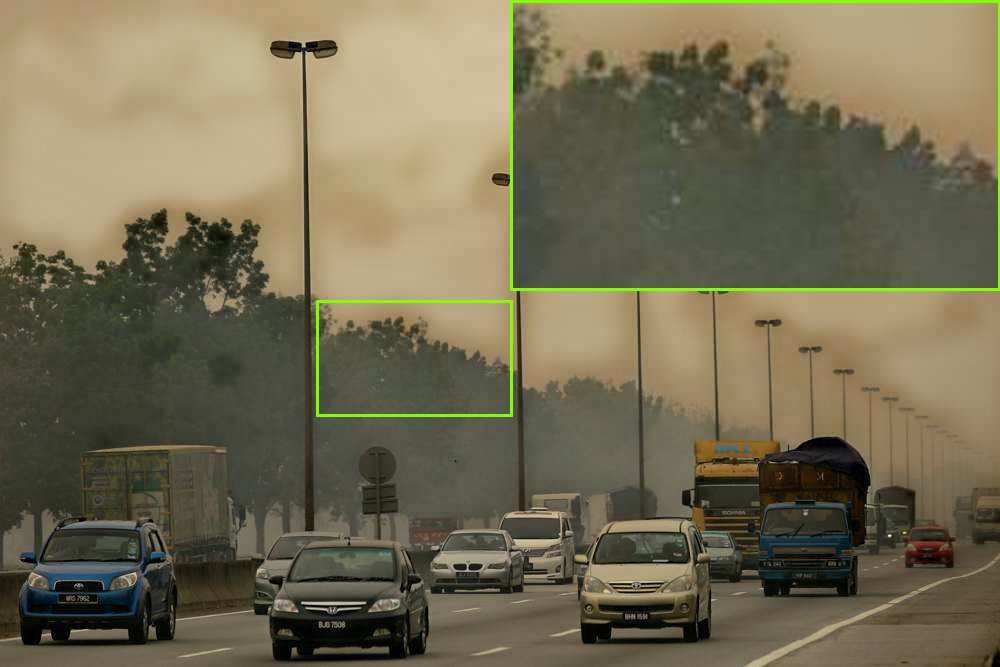}
            \small{(d) D4~\cite{d4}}
	\end{minipage}
	\begin{minipage}[h]{0.138\linewidth}
		\centering
		\includegraphics[width=\linewidth]{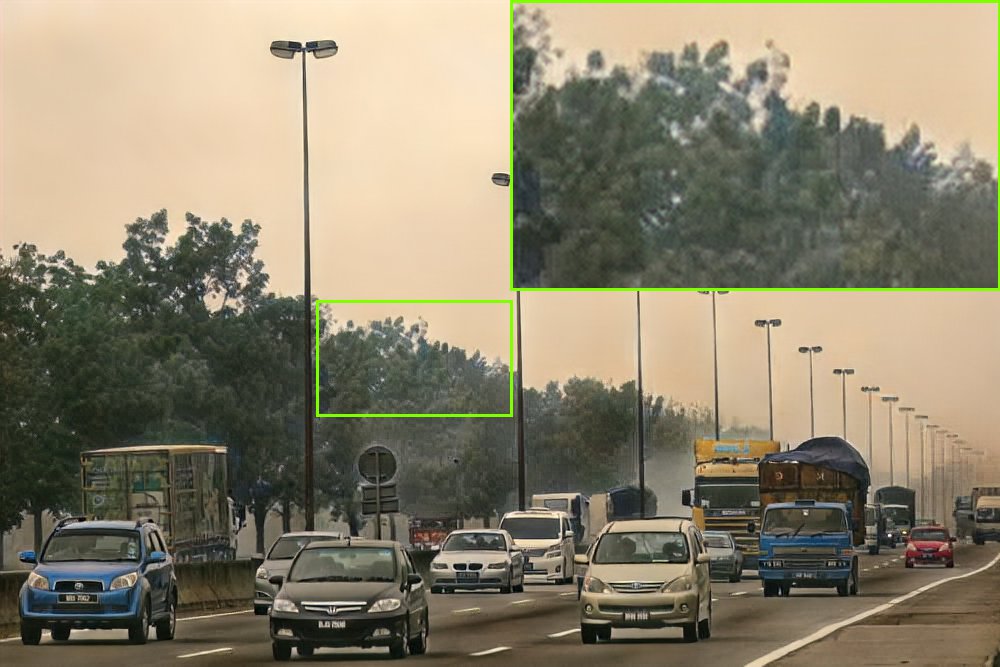}
            \small{(e) RIDCP~\cite{ridcp}}
	\end{minipage}
	\begin{minipage}[h]{0.138\linewidth}
		\centering
		\includegraphics[width=\linewidth]{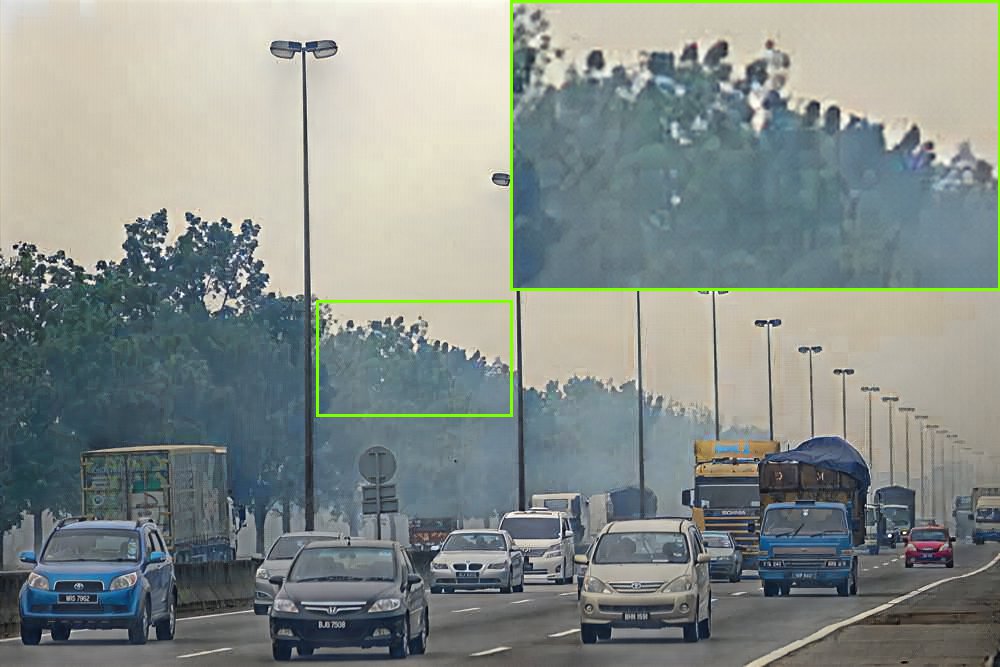}
            \small{(f) PTTD~\cite{pttd}}
	\end{minipage}
	\begin{minipage}[h]{0.138\linewidth}
		\centering
		\includegraphics[width=\linewidth]{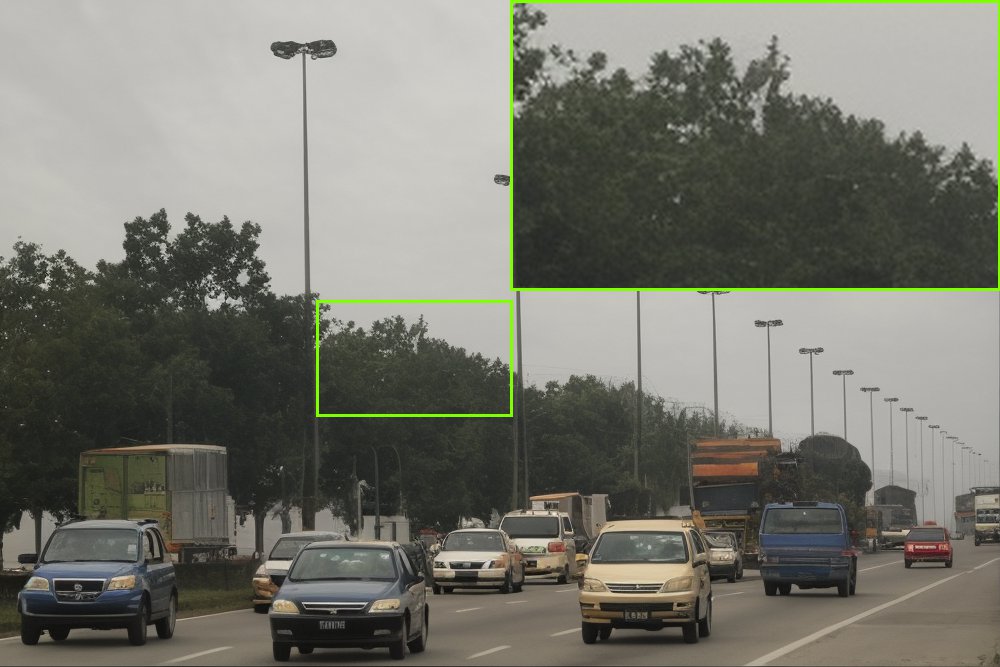}
            \small{(g) Ours}
	\end{minipage}
	\caption{Visual comparisons on the RTTS dataset~\cite{reside}. Zoomed-in for details.} 
	\label{fig:qualitative_RTTS}
\end{figure*}

At timestep $\omega$, the sampling process resumes from $z_\omega$, refining the result and adding further details, especially in densely hazy regions. Compared with initiating sampling from purely random noise at timestep $t = T$, our method provides enhanced sampling fidelity because the dehazing estimate $\hat{y}$ derives content from the input image.

\noindent\textbf{Guided Refinement Stage.} To further improve sampling fidelity, we propose a haze density-aware fidelity guidance mechanism, which guides the denoising process towards the dehazing estimate $\hat{y}$ during refinement. In general, densely hazy regions benefit more from generated contents, while slightly hazy regions should emphasize fidelity to the input image. To achieve this adaptive weighting, we compute a rough transmission map $\mathcal{W}$ for the hazy input image $x$ using the Dark Channel Prior (DCP) algorithm~\cite{prior1_dcp}. This transmission map effectively represents the inverse of haze density because regions with lighter haze exhibit higher transmission values. Consequently, we use $\mathcal{W}$ directly as a weighting map.

At each timestep $t$, we obtain the predicted clean image $\mathcal{D}(\hat{z}_0^{(t)})$ based on Equation~\ref{eq:predict_mean}. We then employ the MS-SSIM loss~\cite{msssim}, which is sensitive to structural similarity, as our fidelity metric. Formally, the fidelity loss is defined as:
\begin{equation}
  \mathcal{L}(\mathcal{D}(\hat{z}_0^{(t)})) = \textit{MS-SSIM}(\mathcal{D}(\hat{z}_0^{(t)}),\; \hat{y}).
  \label{eq:fidelity_loss}
\end{equation}

We apply gradient descent to optimize $\hat{z}_0^{(t)}$ towards higher fidelity. Specifically, the gradient of the fidelity loss with respect to $\mathcal{D}(\hat{z}_0^{(t)})$ is multiplied elementwise with the weighting map $\mathcal{W}$ to selectively reduce guidance strength in densely hazy regions. At each refinement step, $\hat{z}_0^{(t)}$ is updated according to:
\begin{equation}
  \hat{z}_0^{(t)}\leftarrow\hat{z}_0^{(t)}-s\cdot \mathit{J}_\mathcal{D}^T(\hat{z}_0^{(t)})(\mathcal{W}\otimes\nabla_{\mathcal{D}(\hat{z}_0^{(t)})}\mathcal{L}(\mathcal{D}(\hat{z}_0^{(t)}))),
  \label{eq:fidelity_loss}
\end{equation}
where $s$ controls the guidance strength, $\mathit{J}_\mathcal{D}(\hat{z}_0^{(t)})$ is the Jacobian matrix of the decoder $\mathcal{D}$, and $\otimes$ denotes elementwise multiplication. In this scheme, the guidance strength $s$ offers a trade-off between generated image \textit{quality} and \textit{fidelity} to the hazy image. Finally, $z_{t-1}$ is computed using the updated $\hat{z}_0^{(t)}$ and $z_t$, after which the sampling process moves to the next timestep.

\section{Experiments}

\subsection{Experiment Settings}

\noindent\textbf{Datasets.}
Around 4,800 real-world hazy images from the URHI split of the RESIDE dataset~\cite{reside}, together with synthetic image pairs generated by the phenomenological degradation pipeline from RIDCP~\cite{ridcp}, were used for the training of HazeGen. Subsequently, we generated 100,000 realistic hazy images with HazeGen to provide high-quality training data for DiffDehaze. For qualitative and quantitative evaluations, we employed the widely used RTTS split from RESIDE~\cite{reside}, which contains 4,322 images covering diverse scenes and haze patterns. The Fattal’s dataset~\cite{prior6_fattal_clp} is used for additional visual comparison, which comprises 31 classical hazy images.

\begin{table*}[t]
	\centering

	\caption{Quantitative comparisons of various dehazing methods on the RTTS dataset~\cite{reside}. \textbf{Bold} numbers indicate the best performance.}
	\adjustbox{width=0.95\linewidth}{
		\begin{tabular}{l|cccccccc}
			\toprule
			Method & Venue & FADE$\downarrow$  & Q-Align$\uparrow$ & LIQE$\uparrow$ & CLIPIQA$\uparrow$ & ManIQA$\uparrow$ & MUSIQ $\uparrow$ & BRISQUE$\downarrow$\\
			\midrule
			\midrule
			Hazy Input & - &2.484 & 2.0586 & 1.9146 & 0.3882 & 0.3081 & 53.768 & 36.6423 \\
			DAD~\cite{dad} & CVPR'20 & 1.130 & 2.0117 & 1.6666 & 0.2512 & 0.2219 & 49.337 & 32.4565 \\
			PSD~\cite{psd} & CVPR'21 & 0.920 & 1.9014 & 1.5174 & 0.2497 & 0.2640 & 52.806 & 21.6160 \\
			D4~\cite{d4} & CVPR'22 & 1.358 & 2.0801 & 1.9741 & 0.3404 & 0.2974 & 53.555 & 28.1015 \\
			RIDCP~\cite{ridcp} & CVPR'23 & 0.944 & 2.4844 & 2.5518 & 0.3367 & 0.2769 & 59.384 & 17.2944 \\
                PTTD~\cite{pttd} & ECCV'24  & \textbf{0.712} & 2.2891 & 2.3532 & 0.3700 & 0.3095 & 62.114 & 16.6302 \\
			\midrule

			DiffDehaze (Ours) & - & 1.138 & \textbf{2.8340} & \textbf{3.0693} & \textbf{0.4263} & \textbf{0.3661} & \textbf{65.086} & \textbf{16.4924}\\
			\bottomrule
		\end{tabular}}
	\label{tab:rtts}
\end{table*}

\begin{figure*}[t]
	\centering
 \begin{minipage}[h]{0.138\linewidth}
		\centering
		\includegraphics[width=\linewidth]{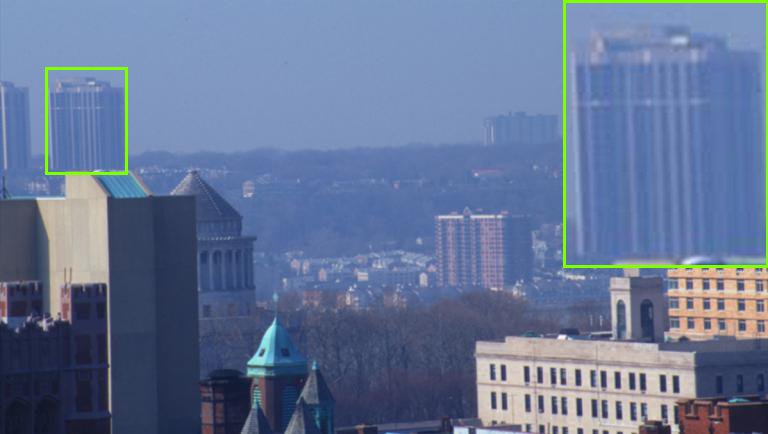}
	\end{minipage}
	\begin{minipage}[h]{0.138\linewidth}
		\centering
		\includegraphics[width=\linewidth]{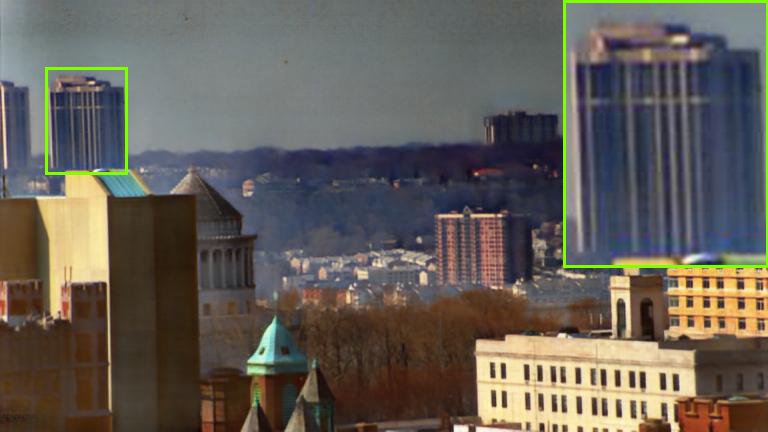}
	\end{minipage}
	\begin{minipage}[h]{0.138\linewidth}
		\centering
		\includegraphics[width=\linewidth]{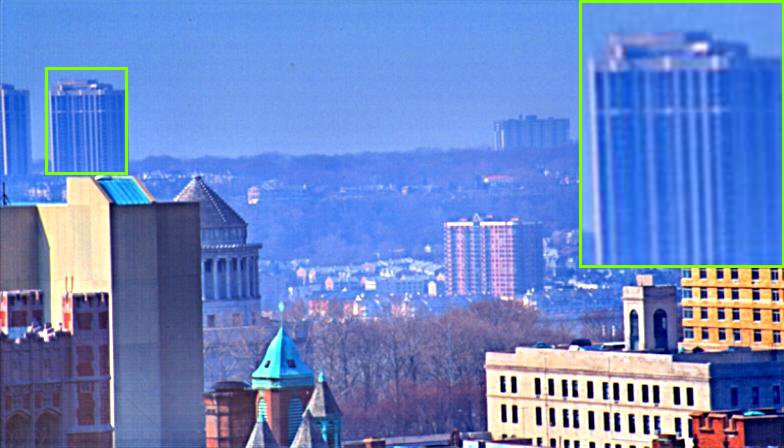}
	\end{minipage}
	\begin{minipage}[h]{0.138\linewidth}
		\centering
		\includegraphics[width=\linewidth]{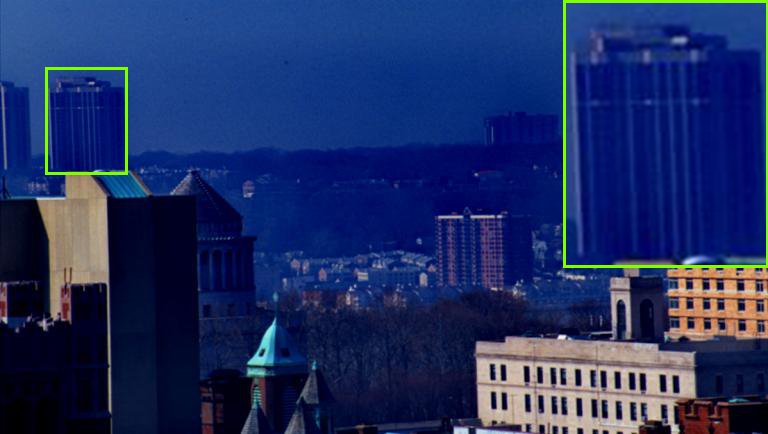}
	\end{minipage}
	\begin{minipage}[h]{0.138\linewidth}
		\centering
		\includegraphics[width=\linewidth]{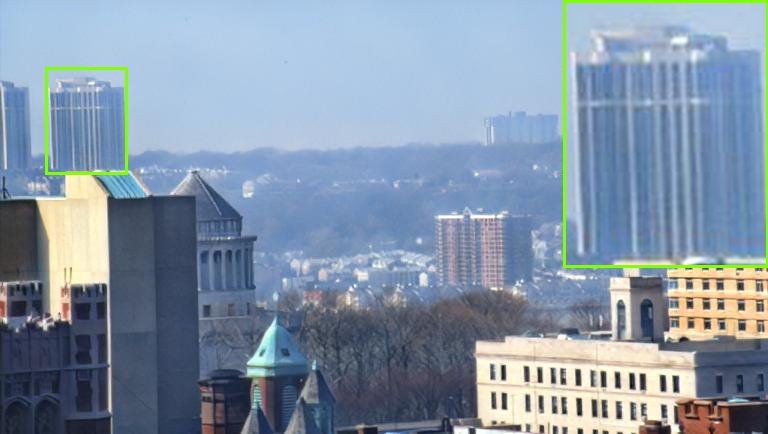}
	\end{minipage}
	\begin{minipage}[h]{0.138\linewidth}
		\centering
		\includegraphics[width=\linewidth]{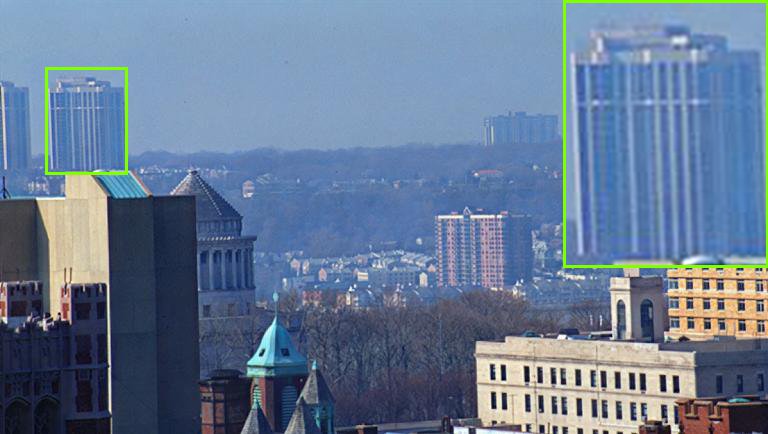}
	\end{minipage}
	\begin{minipage}[h]{0.138\linewidth}
		\centering
		\includegraphics[width=\linewidth]{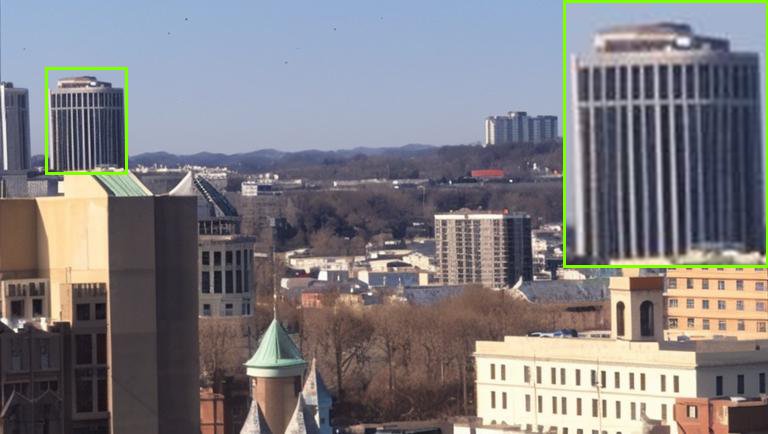}
	\end{minipage}
    \begin{minipage}[h]{0.138\linewidth}
		\centering
		\includegraphics[width=\linewidth]{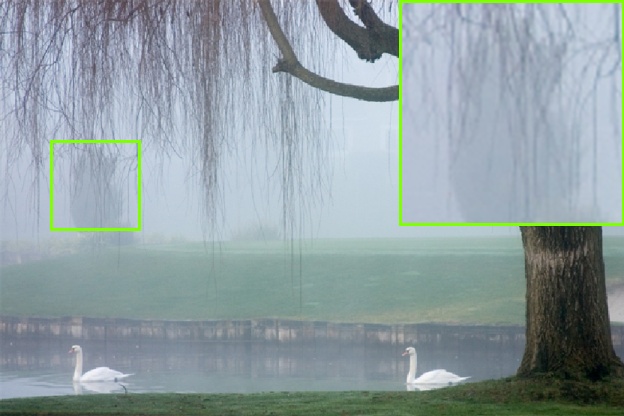}
            \small{(a) Input}
	\end{minipage}
	\begin{minipage}[h]{0.138\linewidth}
		\centering
		\includegraphics[width=\linewidth]{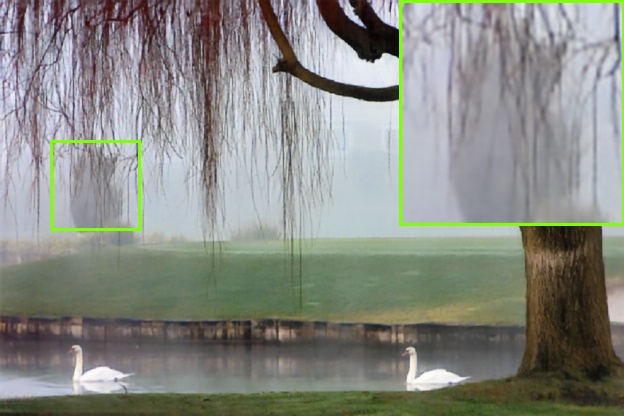}
            \small{(b) DAD~\cite{dad}}
	\end{minipage}
	\begin{minipage}[h]{0.138\linewidth}
		\centering
		\includegraphics[width=\linewidth]{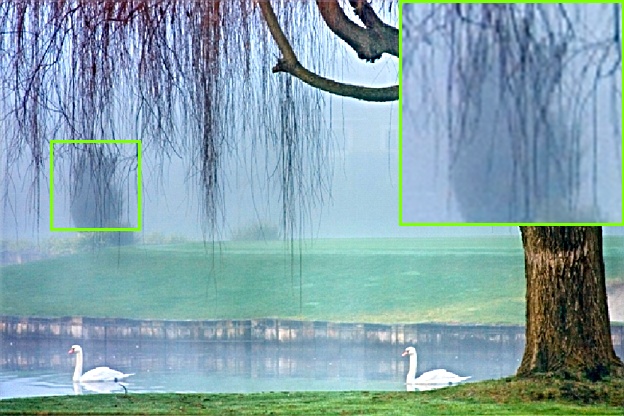}
            \small{(c) PSD~\cite{psd}}
	\end{minipage}
	\begin{minipage}[h]{0.138\linewidth}
		\centering
		\includegraphics[width=\linewidth]{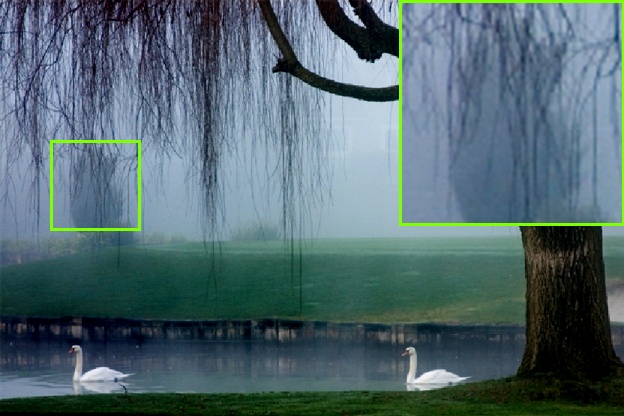}
            \small{(d) D4~\cite{d4}}
	\end{minipage}
	\begin{minipage}[h]{0.138\linewidth}
		\centering
		\includegraphics[width=\linewidth]{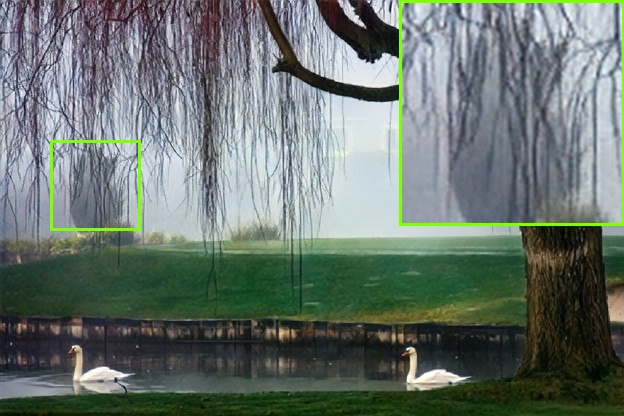}
            \small{(e) RIDCP~\cite{ridcp}}
	\end{minipage}
	\begin{minipage}[h]{0.138\linewidth}
		\centering
		\includegraphics[width=\linewidth]{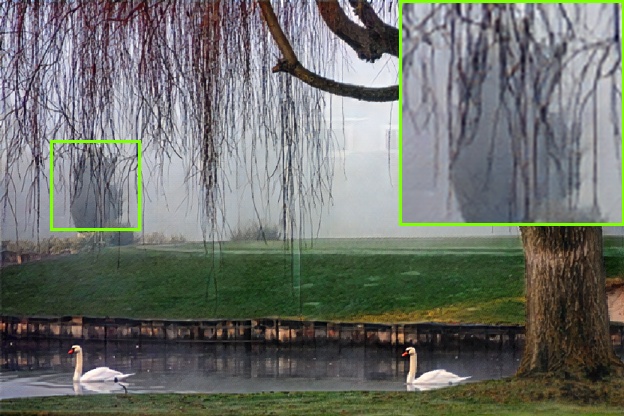}
            \small{(f) PTTD~\cite{pttd}}
	\end{minipage}
	\begin{minipage}[h]{0.138\linewidth}
		\centering
		\includegraphics[width=\linewidth]{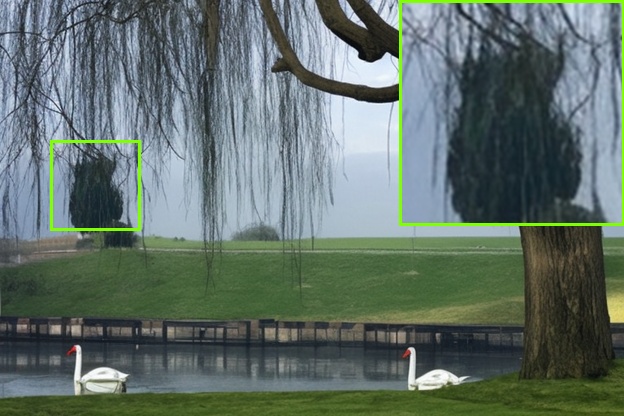}
            \small{(g) Ours}
	\end{minipage}
	\caption{Visual comparisons on the Fattal's dataset~\cite{prior6_fattal_clp}. Zoomed-in for details.}
	\label{fig:qualitative_Fattal}
\end{figure*}

\noindent\textbf{Implementation Details.}
The proposed pipeline is implemented using PyTorch 2.2.2 and is built upon Stable Diffusion v2-1~\cite{stablediffusion}. HazeGen is optimized with the AdamW optimizer~\cite{adamw}, employing a learning rate of $3 \times 10^{-5}$. The probability parameter $p$ is set to 0.3, the training batch size is 16, and training lasts for 5,000 iterations. For the sampling of HazeGen, we adopt the spaced DDPM sampler~\cite{iddpm} with 50 steps and set the mixture coefficient $w$ to 0.85. The training of DiffDehaze adopts the same optimizer and learning rate but lasts for 55,000 iterations. To keep optimal image quality, DiffDehaze employs AccSamp with 50 steps and empirically set parameters $\tau=800$, $\omega=600$, and guidance strength $s=0.1$. It's worth noting that reducing $\omega$ or increasing $s$ still produces satisfactory results.

\subsection{Comparison with State-of-the-Art Methods}
\hspace*{\parindent} We evaluate the proposed method against several state-of-the-art approaches for real-world image dehazing, including DAD~\cite{dad}, PSD~\cite{psd}, D4~\cite{d4}, RIDCP~\cite{ridcp}, and PTTD~\cite{pttd}. Comprehensive experiments are conducted to provide a thorough assessment.

\noindent\textbf{Quantitative Comparisons.}
As real-world hazy datasets lack ground-truth clean images, we adopt multiple no-reference image quality metrics for quantitative evaluation. A detailed evaluation on the RTTS dataset using FADE~\cite{fade}, Q-Align~\cite{qalign}, LIQE~\cite{liqe}, CLIPIQA~\cite{clipiqa}, ManIQA~\cite{maniqa}, MUSIQ~\cite{musiq}, and BRISQUE~\cite{brisque} is presented in Table~\ref{tab:rtts}. Our method achieves leading performance across all metrics except for FADE. Specifically, it attains a remarkable 23.8\% improvement on the Q-Align metric, as well as substantial improvements on other recent metrics such as LIQE and CLIPIQA. 	However, our method has relatively worse performance on FADE, which is partly due to FADE’s unreliability in evaluating dehazing quality. The limitations of FADE will be further discussed in the next section. Overall, our approach demonstrates superior quantitative performance compared to existing methods.

\noindent\textbf{Qualitative Comparisons.}
Qualitative results comparing our method and state-of-the-art real-world dehazing methods on the RTTS dataset~\cite{reside} are illustrated in Figure~\ref{fig:qualitative_RTTS}. Notably, our method is capable of recovering vivid details (e.g., the trees in the fifth row), whereas other methods fail to produce such details. Additional qualitative comparisons using Fattal’s dataset~\cite{prior6_fattal_clp} are shown in Figure~\ref{fig:qualitative_Fattal}. Images generated by DAD, PSD, D4, RIDCP, and PTTD often contain residual haze, especially in heavily hazy areas. In contrast, our method consistently removes haze from all regions, producing visually appealing images with natural color restoration that closely resemble clear weather conditions. 



\subsection{Ablation Studies}



\begin{table}[t]
	\centering

	\caption{Ablation study of HazeGen on the RTTS dataset~\cite{reside}. \textbf{Bold} numbers indicate the best performance.}
	\adjustbox{width=\linewidth}{
		\begin{tabular}{c|ccccc}
			\toprule
			Variant  & Q-Align$\uparrow$ & LIQE$\uparrow$ & CLIPIQA$\uparrow$ & ManIQA$\uparrow$ & MUSIQ $\uparrow$\\
			\midrule
			w\slash o hybrid  & 2.6582 & 2.7754 & 0.4174 & 0.3513 & 63.079 \\
			w\slash o blended  & 2.5410 & 2.6410 & 0.4262 & 0.3510 & 62.489 \\
            w\slash o both & 2.8223 & 1.5983 & 0.3266 & 0.2782 & 49.933 \\
			w\slash o HazeGen & 2.1660 & 2.0014 & 0.3540 & 0.2928 & 55.871 \\
			\midrule

			full version & \textbf{2.8340} & \textbf{3.0693} & \textbf{0.4263} & \textbf{0.3661} & \textbf{65.086}\\
			\bottomrule
		\end{tabular}}
	\label{tab:ablation1}
\end{table}

\begin{figure}[t]
	\centering
    \begin{minipage}[h]{0.135\linewidth}
		\centering
		\includegraphics[width=\linewidth]{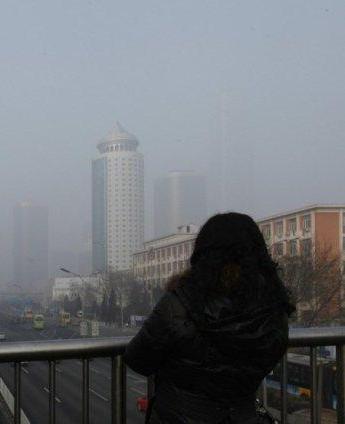}
	\end{minipage}
	\begin{minipage}[h]{0.27\linewidth}
		\centering
		\includegraphics[width=\linewidth]{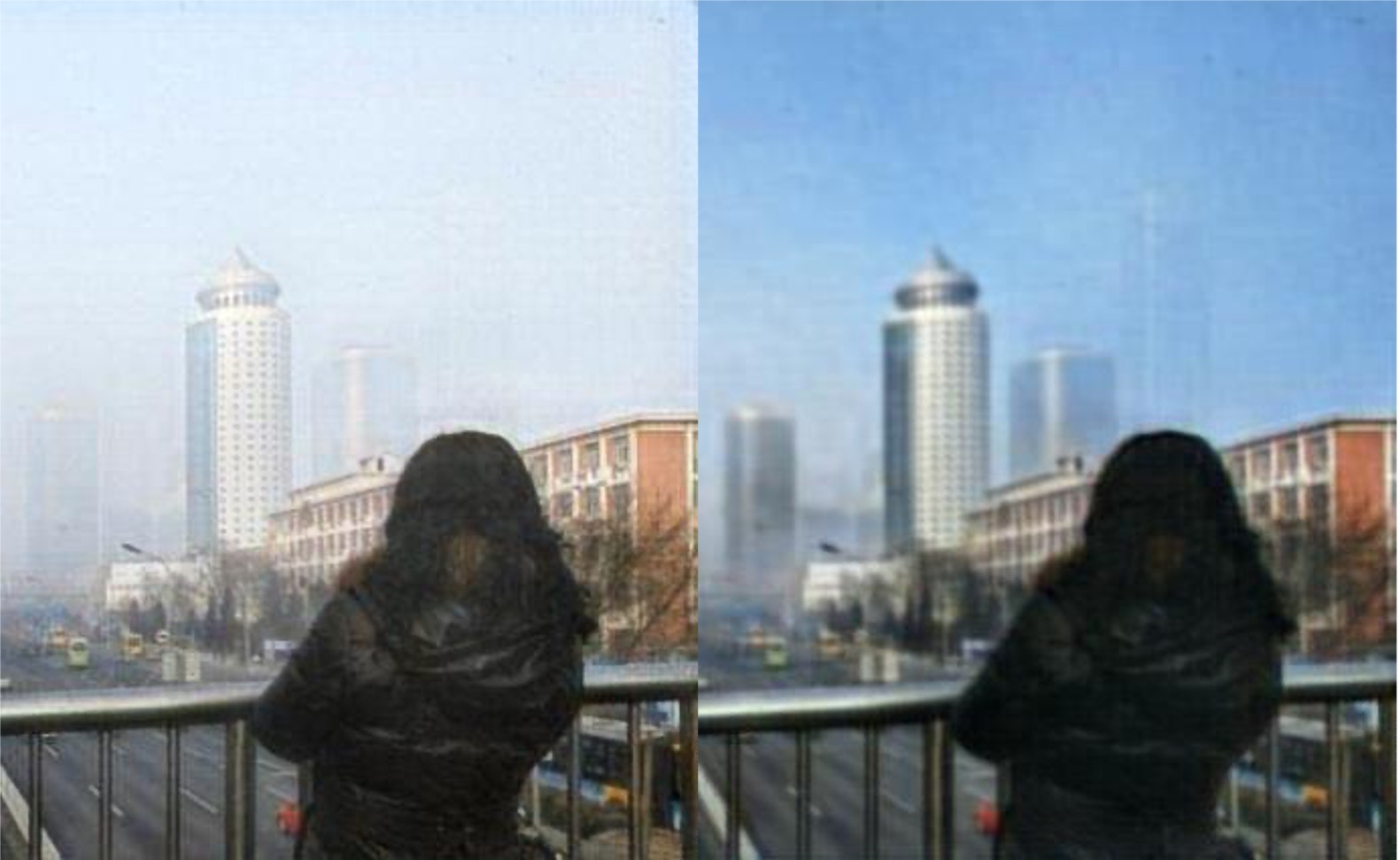}
	\end{minipage}
	\begin{minipage}[h]{0.27\linewidth}
		\centering
		\includegraphics[width=\linewidth]{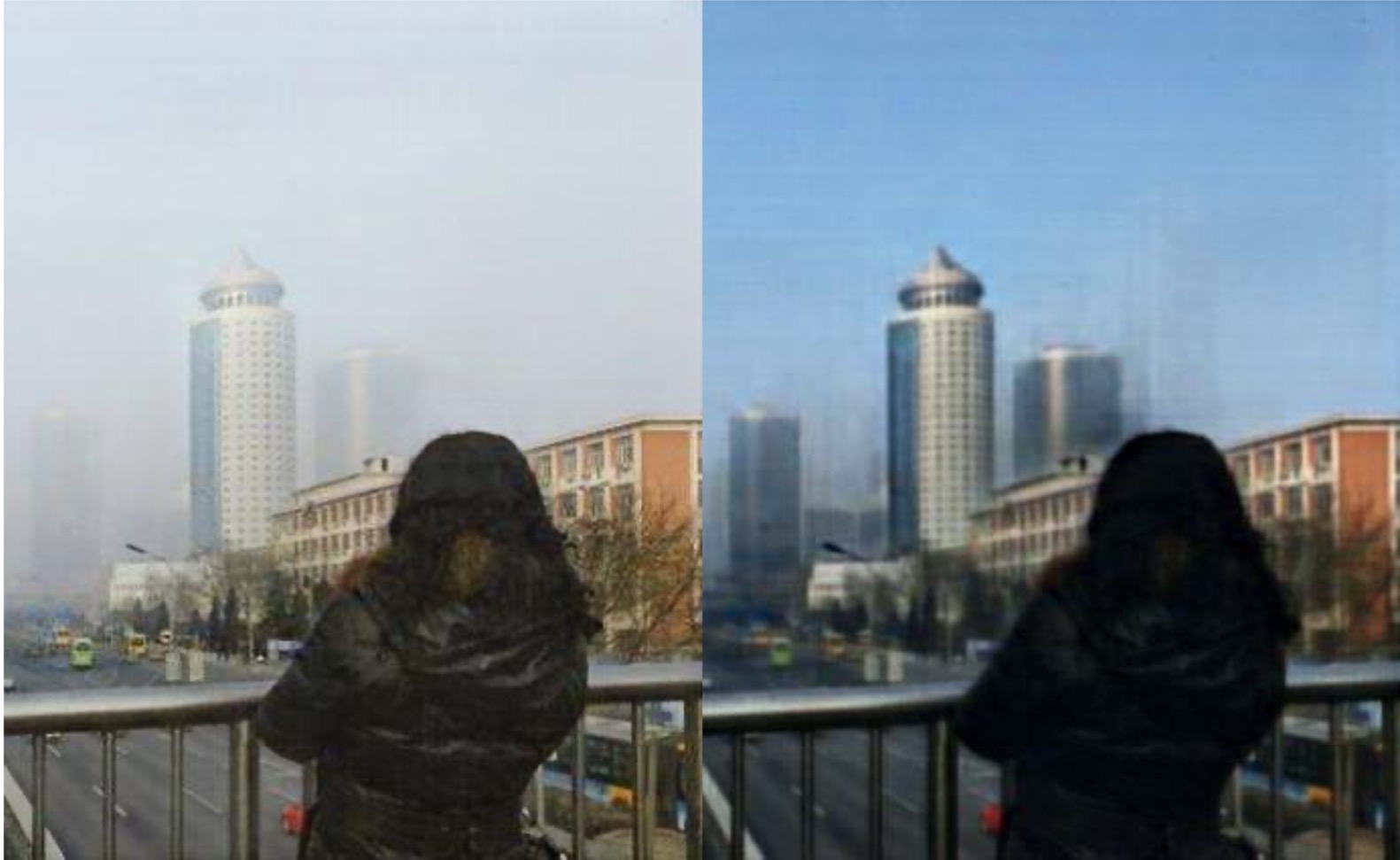}
	\end{minipage}
    	\begin{minipage}[h]{0.27\linewidth}
		\centering
		\includegraphics[width=\linewidth]{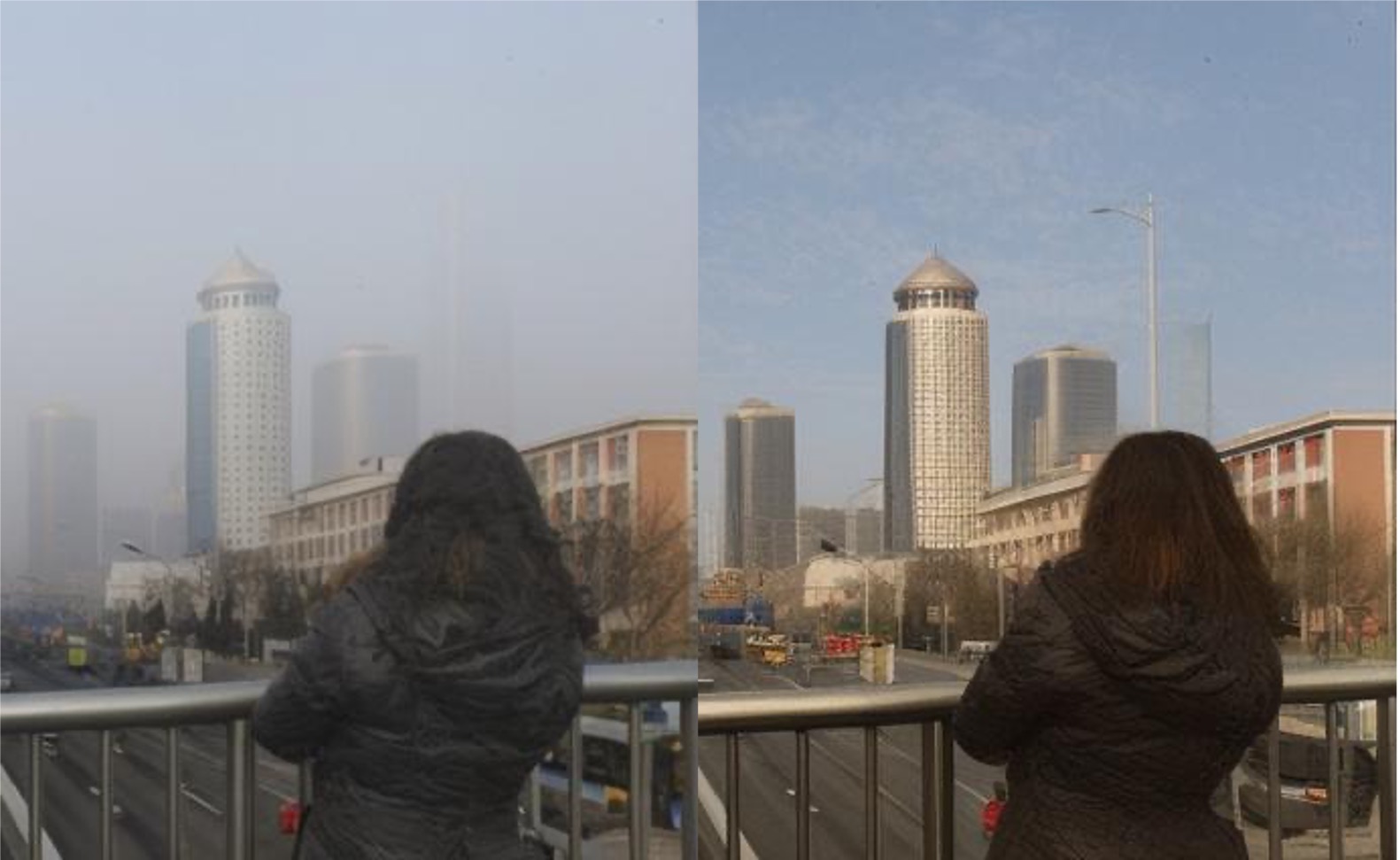}
	\end{minipage}

    \begin{minipage}[h]{0.135\linewidth}
		\centering
		\includegraphics[width=\linewidth]{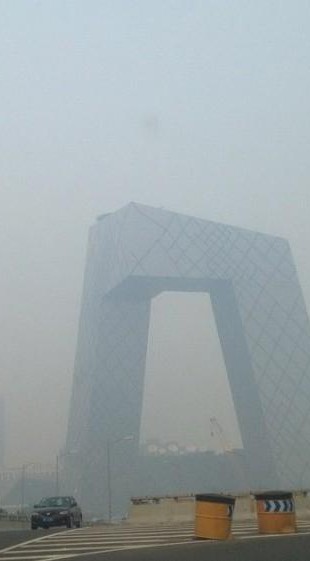}
            \small{(a) Input}
	\end{minipage}
	\begin{minipage}[h]{0.27\linewidth}
		\centering
		\includegraphics[width=\linewidth]{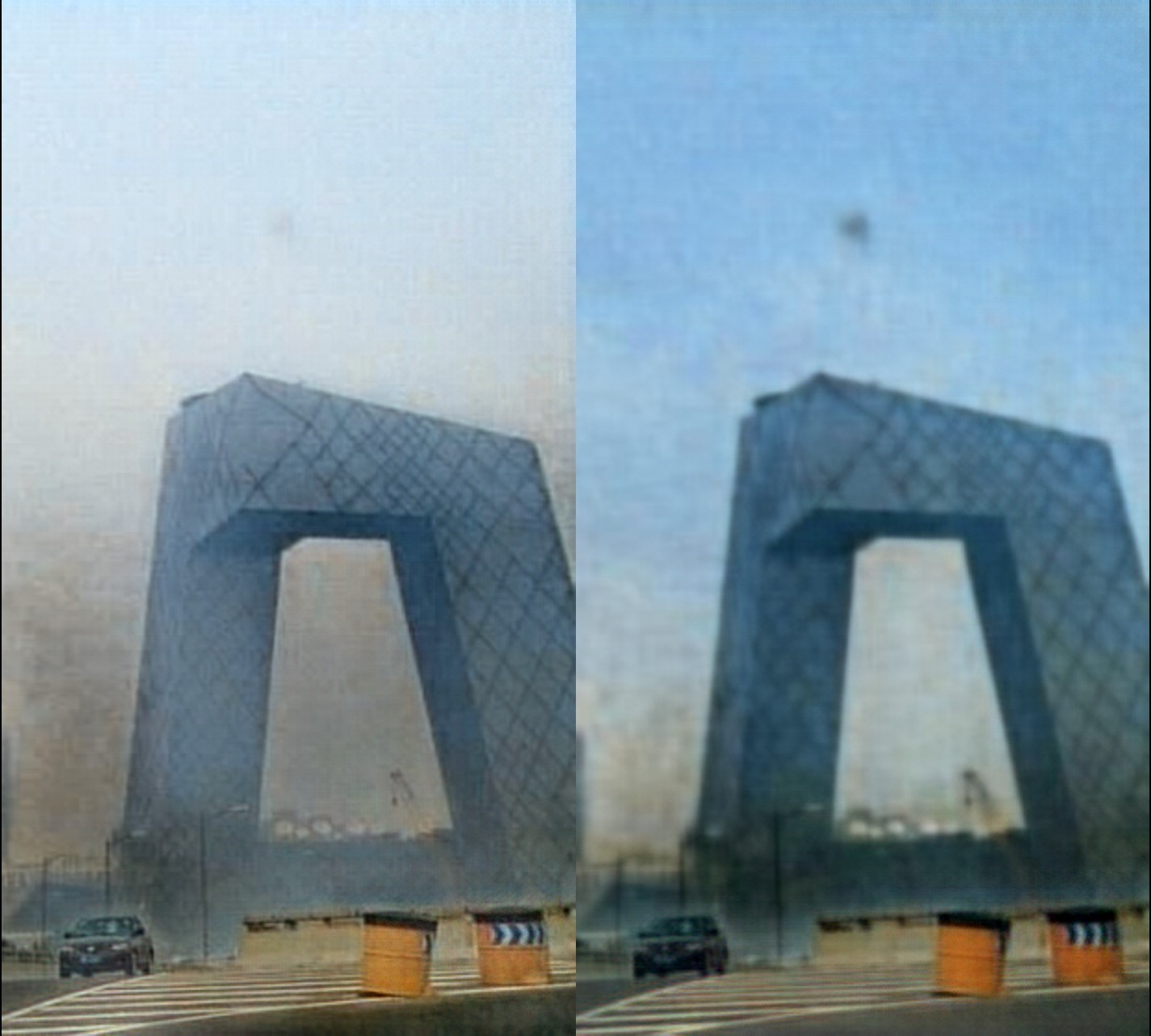}
            \small{(b) MSBDN~\cite{msbdn}}
	\end{minipage}
	\begin{minipage}[h]{0.27\linewidth}
		\centering
		\includegraphics[width=\linewidth]{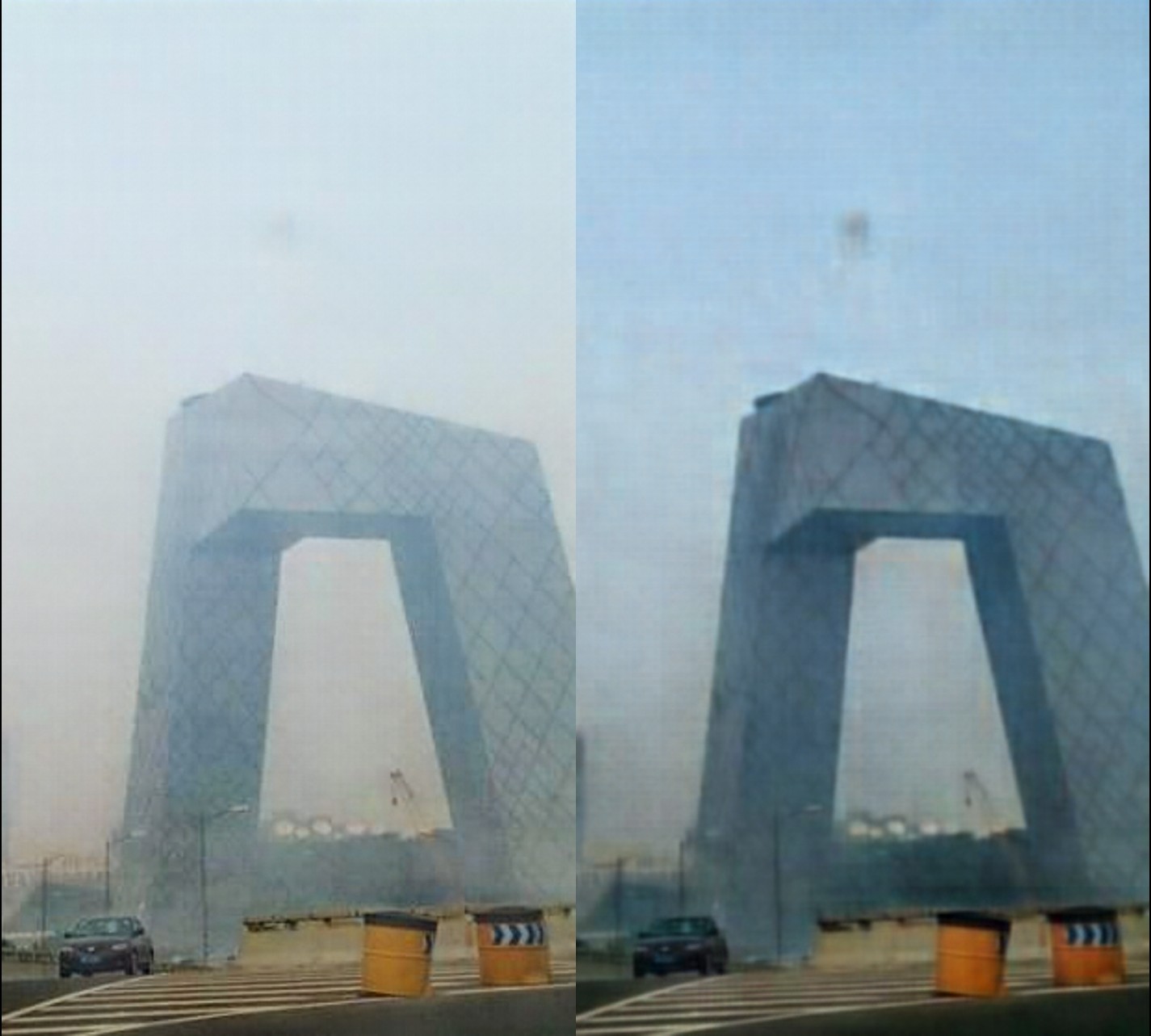}
            \small{(c) NAFNet~\cite{nafnet}}
	\end{minipage}
    \begin{minipage}[h]{0.27\linewidth}
		\centering
		\includegraphics[width=\linewidth]{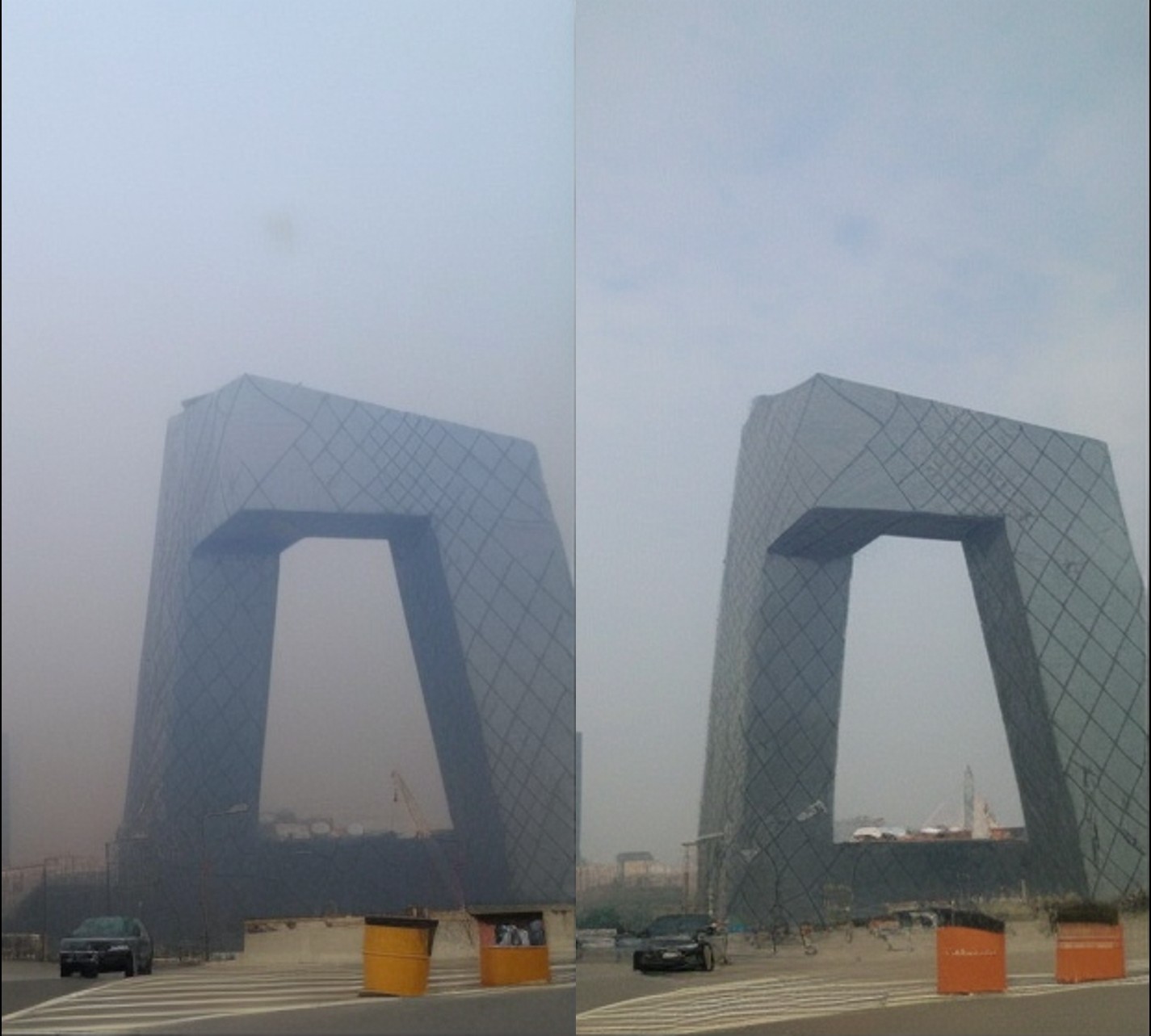}
            \small{(d) Ours}
	\end{minipage}

	\caption{Comparison of training data effectiveness. The left half of each image shows results from models trained on synthetic hazy images from RIDCP~\cite{ridcp}, while the right half shows results from models trained on realistic hazy images generated by HazeGen.}
	\label{fig:ablation1}
\end{figure}
\noindent\textbf{Ablation Study for HazeGen.}
To verify the effectiveness of the proposed training and sampling strategies of HazeGen, we evaluate the performance of DiffDehaze trained on data generated by several variants of HazeGen: (a) without hybrid training (i.e., purely conditional training); (b) without blended sampling; (c) without both hybrid training and blended sampling; and (d) without HazeGen, meaning DiffDehaze is trained directly on synthetic data. The results on the RTTS dataset, presented in Table~\ref{tab:ablation1}, demonstrate that each component is essential to achieve optimal performance. Visual comparisons are provided in the supplementary material.

\noindent\textbf{Effectiveness of HazeGen.}
To further validate the quality of images generated by HazeGen, we compare the dehazing performance of two popular models—MSBDN~\cite{msbdn} and NAFNet~\cite{nafnet}—as well as DiffDehaze, trained separately with synthetic data and data generated by HazeGen. As illustrated in Figure~\ref{fig:ablation1}, all models trained with data generated by HazeGen exhibit improved dehazing capability and overall visual quality.

\begin{figure}[t]
	\centering
    \begin{minipage}[h]{0.31\linewidth}
		\centering
		\includegraphics[width=\linewidth]{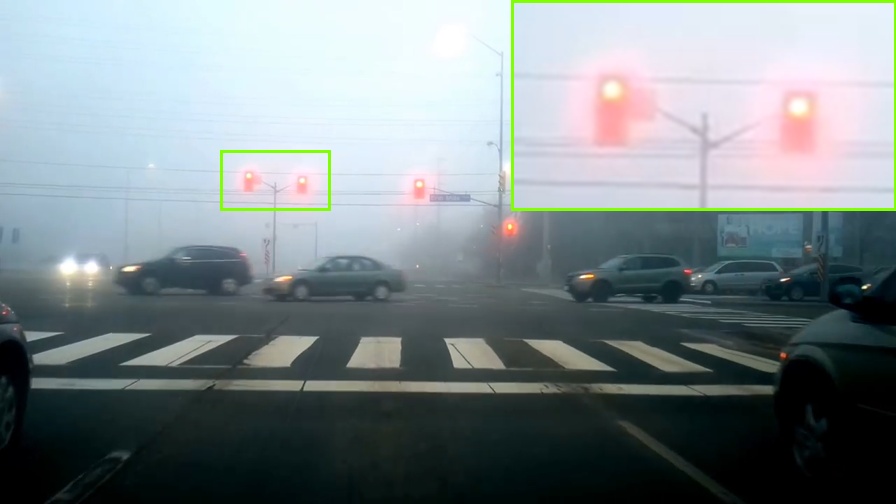}
	\end{minipage}
	\begin{minipage}[h]{0.31\linewidth}
		\centering
		\includegraphics[width=\linewidth]{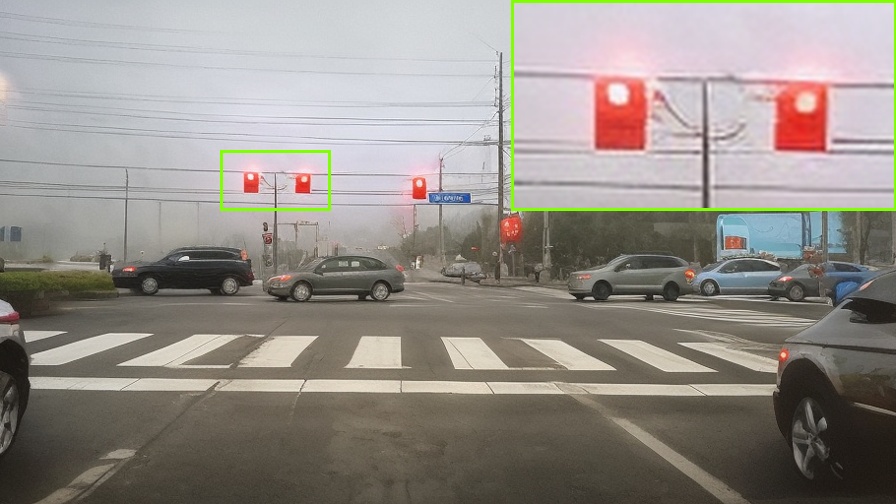}
	\end{minipage}
	\begin{minipage}[h]{0.31\linewidth}
		\centering
		\includegraphics[width=\linewidth]{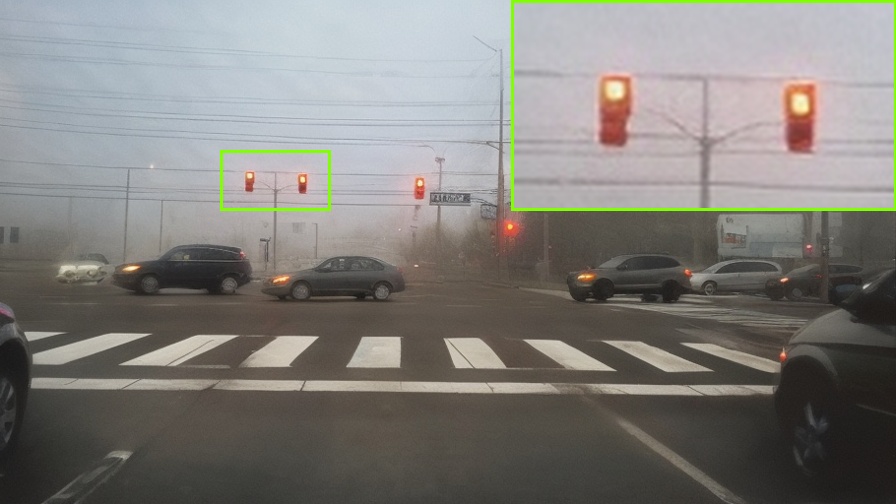}
	\end{minipage}

    \begin{minipage}[h]{0.31\linewidth}
		\centering
		\includegraphics[width=\linewidth]{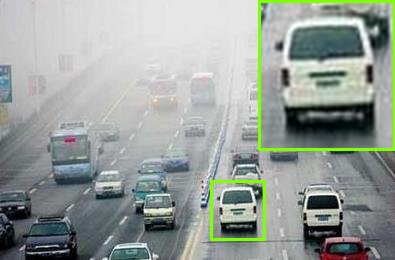}
            \small{(a) Input}
	\end{minipage}
	\begin{minipage}[h]{0.31\linewidth}
		\centering
		\includegraphics[width=\linewidth]{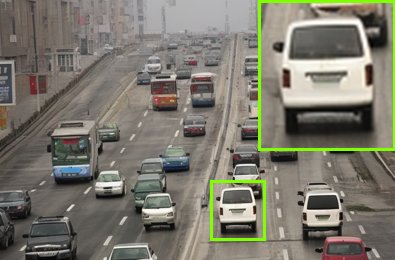}
            \small{(b) Spaced DDPM}
	\end{minipage}
	\begin{minipage}[h]{0.31\linewidth}
		\centering
		\includegraphics[width=\linewidth]{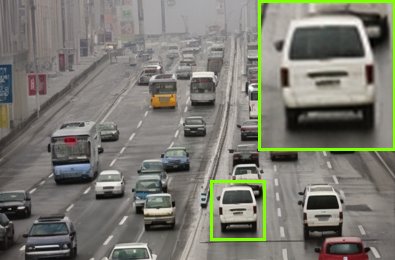}
            \small{(c) AccSamp}
	\end{minipage}

	\caption{Comparison of sampling results with spaced DDPM~\cite{iddpm} and AccSamp samplers.}
	\label{fig:ablation2}
\end{figure}


\begin{table}[t]
	\centering
	\caption{Ablation study of DiffDehaze on the RTTS dataset~\cite{reside}. \textbf{Bold} numbers indicate the best performance.}
	\adjustbox{width=1.0\linewidth}{
		\begin{tabular}{c|ccccc}
			\toprule
			Variant & FADE$\downarrow$ & Q-Align$\uparrow$ & LIQE$\uparrow$ & CLIPIQA$\uparrow$ & ManIQA$\uparrow$ \\
			\midrule
			
            w\slash o AlignOp  & 1.450 & 2.5645 & 2.6786 & 0.4169 & 0.3617\\
            w\slash o weighting  & 1.236 & 2.8067 & 2.9769 & 0.4221 & 0.3457\\
            w\slash o both  & 1.760 & 2.5079 & 2.5837 & 0.4123 & 0.3337\\
            \midrule
            full version  & \textbf{1.138} &  \textbf{2.8340} & \textbf{3.0693} & \textbf{0.4263} & \textbf{0.3661} \\
            \bottomrule
		\end{tabular}}
	\label{tab:ablation}
\end{table}
\noindent\textbf{Ablation Study for DiffDehaze.}
Table~\ref{tab:ablation} reports the ablation results for AccSamp, specifically assessing performance without AlignOp or the haze density weighting mechanism. These results highlight the importance of each individual component. In the absence of AlignOp, the fidelity guidance directly uses the hazy input image for loss computation.

\noindent\textbf{Effectiveness of Fidelity Enhancement.}
Both AlignOp and the adaptive fidelity guidance mechanism enhance the sampling fidelity of AccSamp. To illustrate this improvement, Figure~\ref{fig:ablation2} compares sampling results obtained with the standard spaced DDPM sampler~\cite{iddpm} against AccSamp. The results clearly show that AccSamp significantly increases fidelity to the input images while preserving effective dehazing capability.

\begin{figure}[t]
	\centering
    \begin{minipage}[h]{0.326\linewidth}
		\centering
		\includegraphics[width=\linewidth]{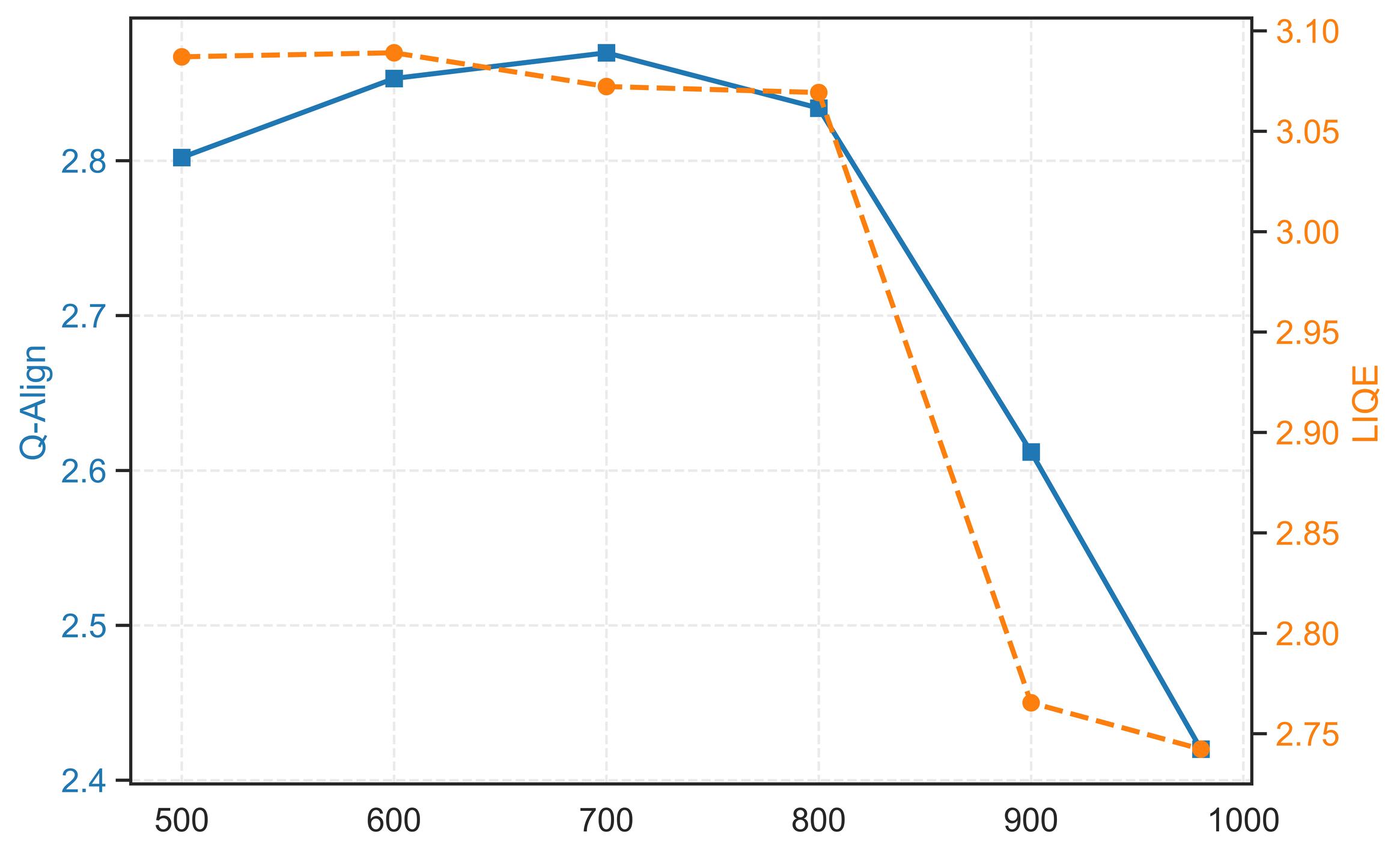}
        \small{$\tau$}
	\end{minipage}
	\begin{minipage}[h]{0.326\linewidth}
		\centering
		\includegraphics[width=\linewidth]{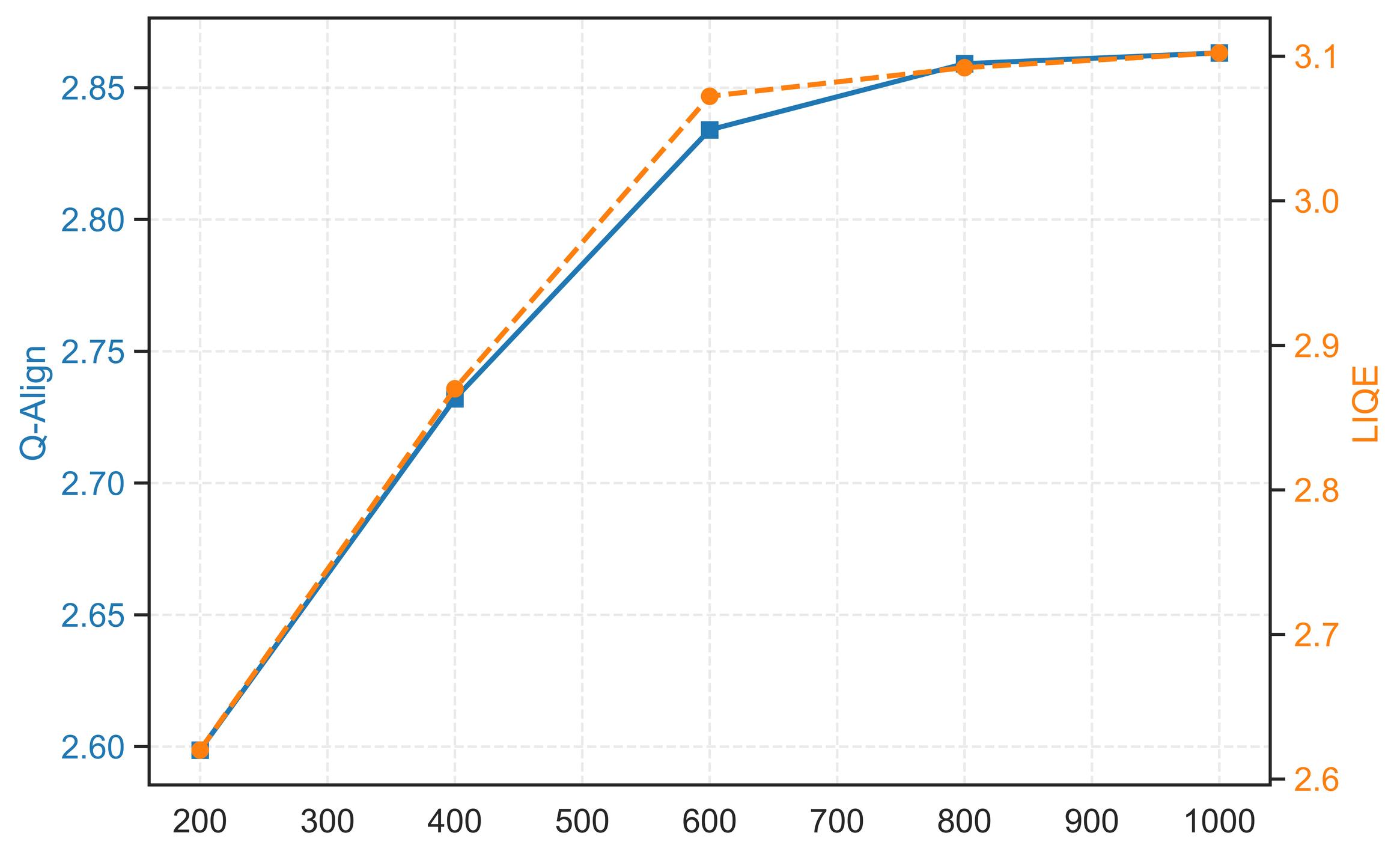}
        \small{$\omega$}
	\end{minipage}
	\begin{minipage}[h]{0.326\linewidth}
		\centering
		\includegraphics[width=\linewidth]{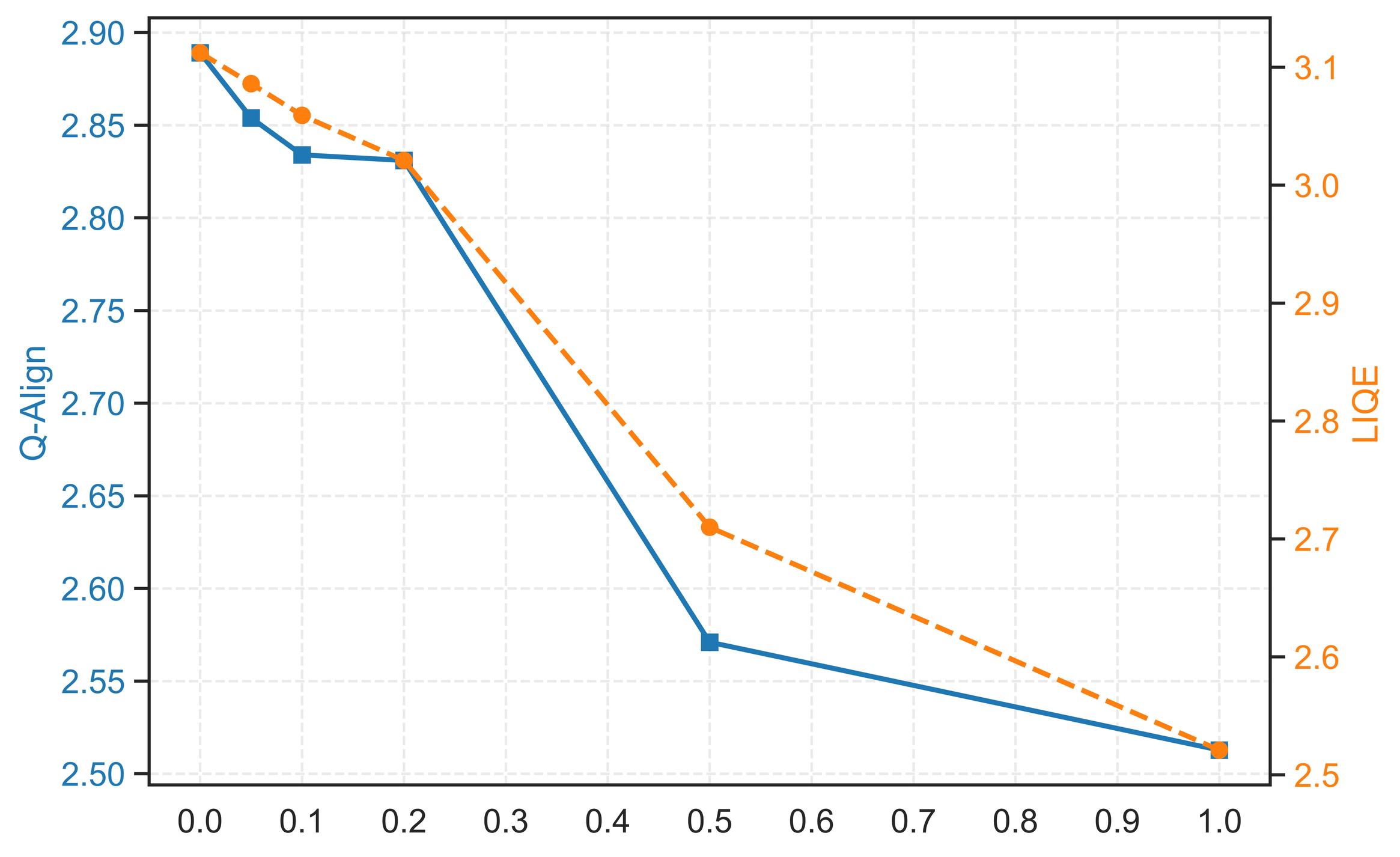}
        \small{$s$}
	\end{minipage}
 \centering
	\caption{Influence of hyperparameters on dehazing performance measured by Q-Align and LIQE.}
	\label{fig:parameter}
\end{figure}
\noindent\textbf{Influence of Hyperparameters in AccSamp.}
Figure~\ref{fig:parameter} presents quantitative metric comparisons (Q-Align and LIQE) across varying hyperparameter settings in AccSamp. Timesteps $\tau$ and $\omega$ are optimized to enhance sampling efficiency without compromising performance, while the guidance strength $s$ balances the trade-off between \textit{quality} and \textit{fidelity}. Therefore, the default hyperparameter configuration is chosen as $\tau=800$, $\omega=600$, and $s=0.1$.

\begin{figure}[t]
	\centering
    \begin{minipage}[h]{0.24\linewidth}
		\centering
		\includegraphics[width=\linewidth]{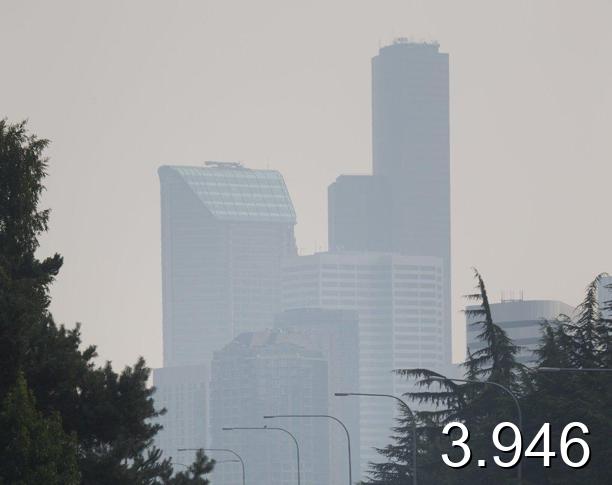}
	\end{minipage}
	\begin{minipage}[h]{0.24\linewidth}
		\centering
		\includegraphics[width=\linewidth]{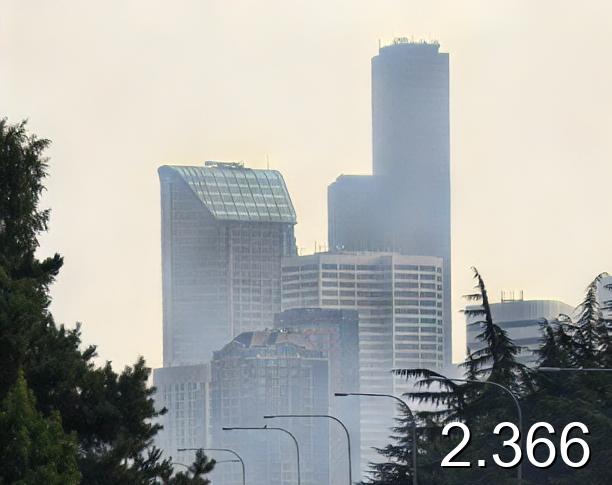}
	\end{minipage}
	\begin{minipage}[h]{0.24\linewidth}
		\centering
		\includegraphics[width=\linewidth]{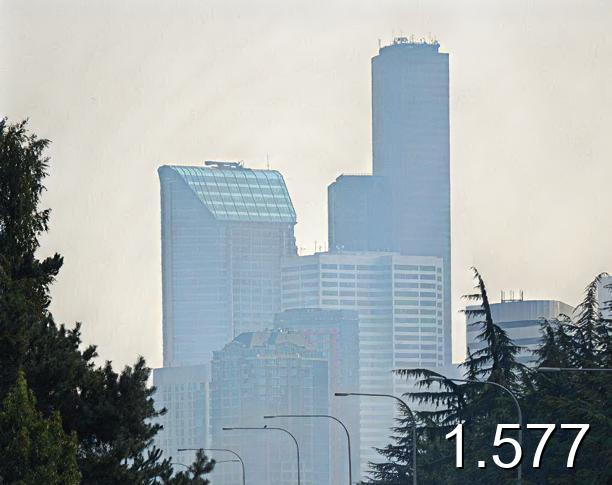}
	\end{minipage}
    	\begin{minipage}[h]{0.24\linewidth}
		\centering
		\includegraphics[width=\linewidth]{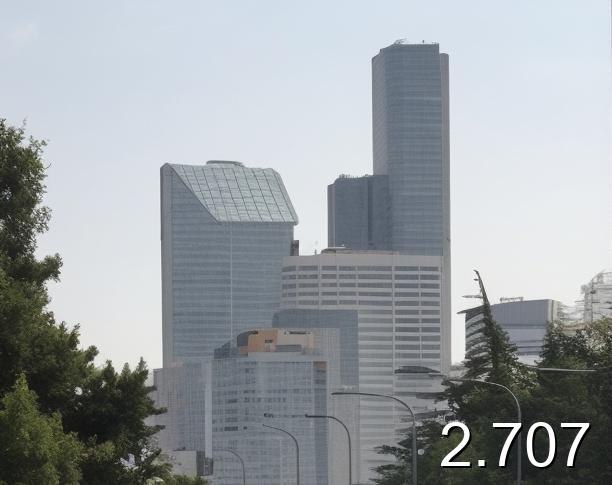}
	\end{minipage}

    \begin{minipage}[h]{0.24\linewidth}
		\centering
		\includegraphics[width=\linewidth]{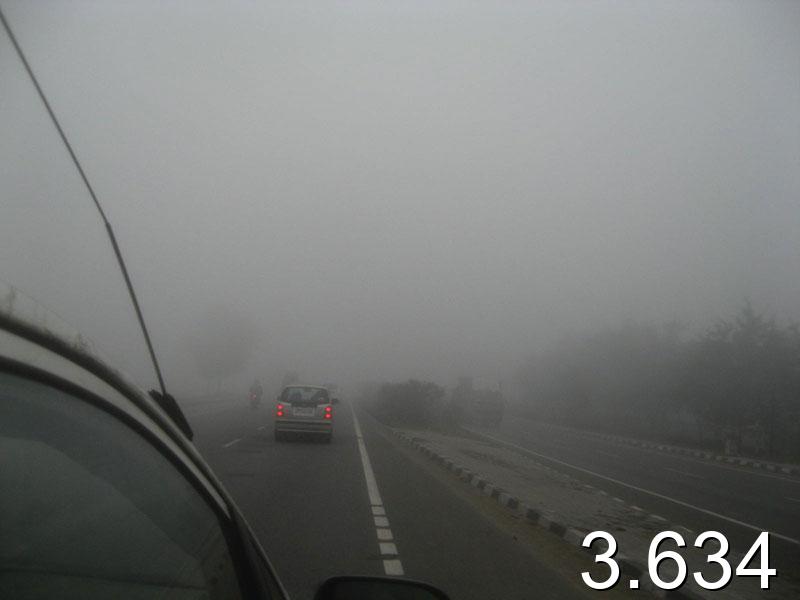}
            \small{(a) Input}
	\end{minipage}
	\begin{minipage}[h]{0.24\linewidth}
		\centering
		\includegraphics[width=\linewidth]{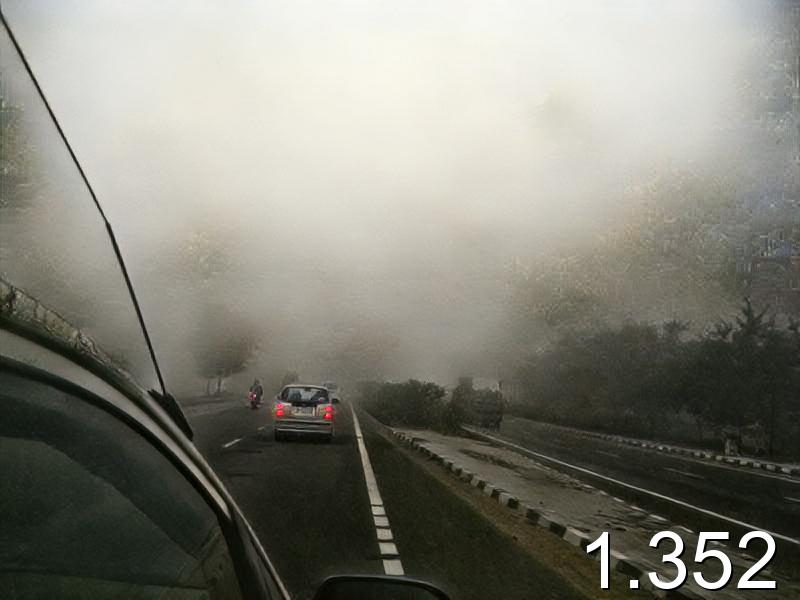}
            \small{(b) RIDCP~\cite{ridcp}}
	\end{minipage}
	\begin{minipage}[h]{0.24\linewidth}
		\centering
		\includegraphics[width=\linewidth]{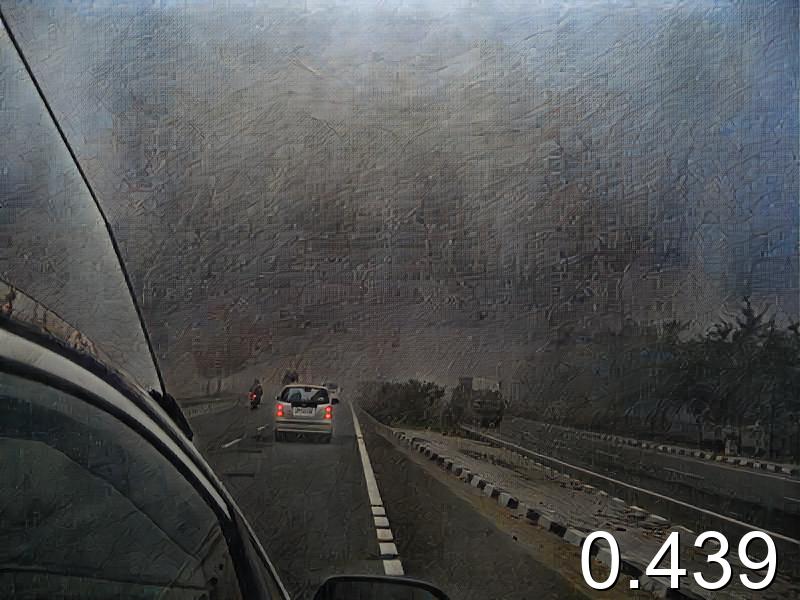}
            \small{(c) PTTD~\cite{pttd}}
	\end{minipage}
    \begin{minipage}[h]{0.24\linewidth}
		\centering
		\includegraphics[width=\linewidth]{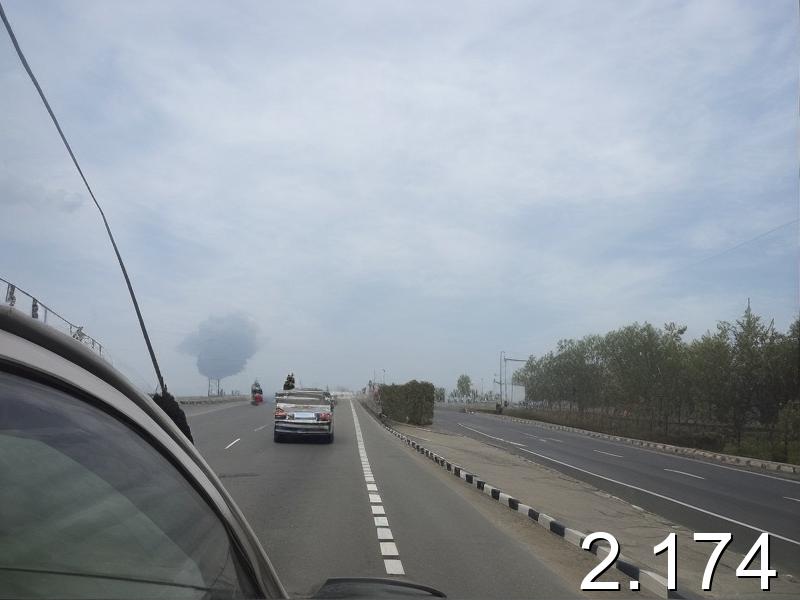}
            \small{(d) Ours}
	\end{minipage}

	\caption{Typical failure cases of FADE evaluation~\cite{fade}. FADE scores are shown at the bottom-right corner of each image.
}
	\label{fig:fade}
\end{figure}
\noindent\textbf{Dehazing Evaluation with FADE~\cite{fade}.} 
Although our method achieves superior dehazing capability, it shows comparatively lower performance in terms of FADE. This discrepancy primarily arises because FADE is insensitive to color-distorted haze residuals. An experiment comparing FADE evaluation results between our method and two state-of-the-art methods is illustrated in Figure~\ref{fig:fade}.  In the first row, RIDCP and PTTD merely shift the haze color toward blue without effective haze removal. In the second row, their results appear visually messy and unclear. However, FADE incorrectly scores these flawed outputs better than ours, contradicting human visual perception.

\vspace{-1mm}
\section{Conclusion}
\vspace{-1mm}
\hspace*{\parindent}In this work, we introduce HazeGen, a novel framework for realistic hazy image generation, and DiffDehaze, a diffusion-based dehazing framework, building upon controlled Stable Diffusion~\cite{stablediffusion} models. By effectively exploiting generative diffusion priors of natural hazy images and employing hybrid training and blended sampling strategies, HazeGen is capable of generating realistic and high-quality hazy images for the training of DiffDehaze. Moreover, leveraging fast dehazing estimates provided by AlignOp, AccSamp can reduce sampling steps and enhance fidelity for DiffDehaze. Comprehensive experiments demonstrated the superior performance of our approach.

\noindent\textbf{Limitations.} Although the proposed method achieves remarkable performance, we identify two notable limitations. First, there remains an urgent need for the development of more sophisticated and reliable metrics for the evaluation of dehazing results. Second, while AccSamp effectively improves sampling fidelity, there is still scope for further enhancement in future research.

\vspace{-2mm}
\section*{Acknowledgment}
\vspace{-1mm}
The work was supported in part by the National Natural Science Foundation of China under Grant 62301310, and in part by Sichuan Science and Technology Program under Grant 2024NSFSC1426.
\vspace{-8mm}
{
    \small
    \bibliographystyle{ieeenat_fullname}
    \bibliography{main}
}


\end{document}